\documentclass[11pt, a4paper, onecolumn, thu]{thuc3i}
\usepackage[authoryear, sort&compress, square]{natbib}
\usepackage{mathtools}
\usepackage{multirow}
\usepackage{xspace}
\usepackage{afterpage}
\usepackage{placeins}
\usepackage{flafter}
\usepackage[most]{tcolorbox}

\newcommand{\sys}{\textsc{CauProphet}\xspace}
\newcommand{\pad}{Causal Belief\xspace}
\newcommand{\cog}{Causal Cognition\xspace}
\newcommand{\scout}{Causal Scout\xspace}

\definecolor{PaperBlue}{HTML}{2A78D6}
\definecolor{PaperGrayDark}{HTML}{5F5F5A}
\definecolor{PaperGray}{HTML}{8A8A85}
\definecolor{PaperGrayLight}{HTML}{BDBDB7}
\definecolor{PaperGrid}{HTML}{E6E6E3}
\definecolor{PaperTableBlue}{HTML}{5A93D0}
\definecolor{PaperTableBlueLight}{HTML}{D3E3F3}
\definecolor{PaperTableBand}{HTML}{DCEAF8}
\definecolor{PaperTableSeparation}{HTML}{DCE8F3}
\definecolor{PaperTableRow}{HTML}{EEEEEA}
\definecolor{PaperMuted}{HTML}{7A8187}
\definecolor{PaperMutedLight}{HTML}{9A9A94}
\definecolor{PaperAbstractBg}{HTML}{F4F7FB}

\colorlet{TableFocus}{PaperTableRow}
\colorlet{TableStripe}{PaperTableRow!45}
\definecolor{TableBest}{HTML}{B4CFE8}
\definecolor{TableSecond}{HTML}{D5E4F2}
\definecolor{TableThird}{HTML}{EEF4FA}

\newcommand{\bestcell}[1]{\cellcolor{TableBest}\textbf{#1}}
\newcommand{\secondcell}[1]{\cellcolor{TableSecond}#1}
\newcommand{\thirdcell}[1]{\cellcolor{TableThird}#1}

\definecolor{AbstractBgColor}{HTML}{F4F7FB}
\renewcommand{\absfont}{\linespread{1.1}\fontsize{10}{12}\selectfont}

\let\reportbibsection\bibsection
\renewcommand{\bibsection}{%
  \reportbibsection
  \phantomsection
  \addcontentsline{toc}{section}{\refname}%
}

\newcommand{\shorttitle}{Teaching Agents to Develop Causal Cognition like Humans}
\title{\sys : Teaching Agents to Develop Causal Cognition like Humans}
\author{%
  Kaiwen Wu$^{1,2}$, Chaochao Lu$^{1,\dagger}$
  \vspace{1mm} \\
  $^1$Shanghai AI Laboratory \quad $^2$Shanghai Jiao Tong University\\
  \texttt{\{wukaiwen,luchaochao\}@pjlab.org.cn}
  \vspace{1mm} \\
  \textbf{$^\dagger$ Corresponding author.}
}
\reportnumber{}
\setheadertext{}
\setheaderlogos{}

\begin{abstract}
Large language model (LLM) agents increasingly support long-horizon tasks by leveraging external memory and past experience. However, existing agents mainly retrieve records or summaries, lacking the ability to form and refine causal understanding of the world. Inspired by human causal cognition, we introduce \sys, an agent that develops and updates causal knowledge within its \cog. For each task, \sys combines relevant knowledge in \cog with current evidence to construct a \pad, a structured belief capturing causal factors and the effects of possible actions. It uses this belief for decision-making and continuously updates both the belief and \cog as new observations emerge, enabling causal knowledge that persists beyond individual experiences. Evaluations demonstrate that \sys progressively acquires causal knowledge and transfers it to future decision-making.
In financial research, \sys leverages prior discoveries to improve downstream analyses and outperforms all baselines in mean rank correlation with next-day returns, achieving a 25.4\% improvement over a baseline agent. It also improves prediction accuracy from 21.3\% to 28.6\% in dynamic real-world settings. Further experiments validate its ability to perform reliable causal analysis and action selection across scientific and everyday tasks, demonstrating the potential of causal experience learning for next-generation LLM agents.
The project repository is available at \url{https://github.com/OpenCausaLab/CauProphet}.
\end{abstract}

\begin{document}
\begin{tcolorbox}[
  colback=AbstractBgColor,
  colframe=AbstractBgColor,
  arc=5pt,
  auto outer arc,
  boxrule=0pt,
  left=8pt, right=8pt,
  top=4pt, bottom=4pt,
  parbox=false,
  width=\textwidth,
  before skip=0pt,
  after skip=0pt,
  enlarge top by=-12pt
]
  \setlength{\parskip}{3pt}
  \maketitle
\end{tcolorbox}
\begin{figure}[!ht]
  \centering
  \includegraphics[width=0.98\linewidth]{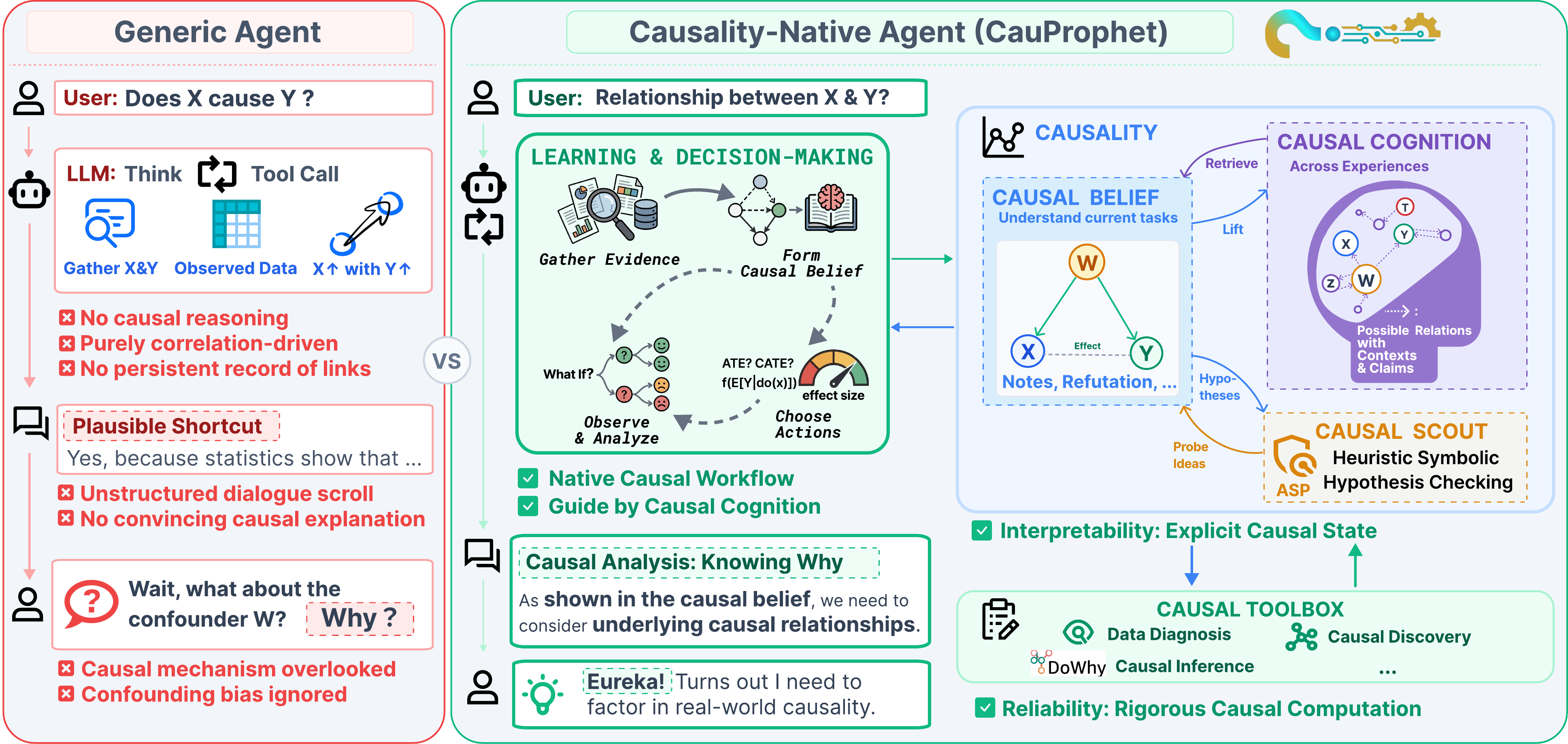}
  \caption{\textbf{Overview of \sys.} \sys maintains explicit causal hypotheses and uses causal skills and tools to test and revise them, keeping its reasoning traceable to evidence and assumptions.}
  \label{fig:teaser}
\end{figure}

\clearpage
\begingroup
  \small
  \setlength{\parskip}{0pt}
  \setcounter{tocdepth}{2}
  \tableofcontents
\endgroup

\clearpage
\section{Introduction}

Large language models (LLMs) have enabled a new class of agents capable of perceiving environments, acting through tools, and pursuing goals~\citep{wang2024survey,yao2023react,schick2023toolformer}.  Within this space, a particularly influential category has emerged in the form of general-purpose autonomous assistants, including Claude Code~\citep{anthropic2025claudecode}, Codex~\citep{openaiCodex}, Hermes~\citep{nousresearch2026hermes}, and OpenClaw~\citep{steinberger2026openclaw}, which interact with users through multi-turn dialogue to address requests involving data analysis, writing, coding, and research. Much of what makes these assistants capable lies not in the base model alone but in the surrounding system---tool interfaces, context management, and safety constraints---that governs the model's behavior through repeated cycles of reasoning, tool use, and observation~\citep{yao2023react,openai2026harness,ning2026code}.

Causality investigates how entities influence one another beyond superficial correlations. It provides the conceptual and mathematical foundation for asking why observed patterns arise, what would happen under an intervention, and what would have occurred under an alternative condition~\citep{pearl2009causality,peters2017elements}. Causal analysis involves identifying the factors that influence an outcome, proposing explanations of how they do so, and examining the evidence and assumptions behind these explanations~\citep{spirtes2000causation,hernan2020causal,dowhy,dowhy-gcm}. Users may need such analysis for their tasks even when they lack expertise in causality. When choosing actions to achieve user goals, LLM agents need to distinguish observed associations from the changes their actions would cause~\citep{pearl2009causality,kiciman2024causal}. Without this distinction, agents may provide plausible but unreliable explanations of why something happens.

Recent work on LLMs and causality follows two main directions: constructing causal knowledge and using it to guide agent behavior. The first automates causal analysis or develops methods for identifying causal relationships and estimating effects~\citep{roy2025causal,peng2026causcientist,peng2026caution,peng2026causcale,zhang2026causight,wang2025CausalCopilot,verma2025cais,wang2026inferenceevolve}; applying its findings to later tasks also requires retaining what was learned and when it applies. The second uses causal knowledge to guide decisions, multimodal reasoning, and program improvement~\citep{yu2025adam,chen2025causalaware,nguyen2025causalplan,chen2024more,chen2026causalevolve}, with reasoning organized around specific environments, multimodal tasks, or programs. These approaches leave open how agents can accumulate causal understanding across experiences as human beings do. Children learn causal relationships through observation and action and use this knowledge to predict outcomes in new situations~\citep{gopnik2004causallearning}. More broadly, human beings draw on an understanding of how events influence one another~\citep{sloman2015causality} to guide decisions without recalling every experience that shaped it.

Inspired by human causal cognition, we introduce \sys (Figure~\ref{fig:teaser}), an LLM agent that accumulates and refines causal knowledge through experience. Within \sys, \cog organizes what the agent learns around recurring factors, how they influence outcomes, and the conditions under which these relationships hold. For a new task, the agent checks which retained explanations fit the current circumstances and combines them with current evidence to form a \pad. This structured belief describes what influences the outcome and how possible actions could change it, guiding the agent's decisions. The same cognition can therefore inform different beliefs as the task and its evidence change. As observations arrive, the agent revises its current belief and updates \cog with selected findings that may inform later decisions. Through this cycle, causal understanding remains useful even when the original experiences are no longer available.

Our experiments demonstrate that \sys progressively develops causal knowledge from experience and applies it to improve decisions in subsequent tasks. We first test whether knowledge retained from earlier tasks improves subsequent financial research, with each task beginning in a fresh context. \sys outperforms all baselines in mean correlation between predicted and actual rankings of next-day stock returns, with a 25.4\% relative gain over an agent with text memory. To test how causal understanding supports judgments across domains as new evidence arrives, we also evaluate predictions made while real-world events unfold. Prediction accuracy increases from 21.3\% for a baseline agent to 28.6\% for \sys. Further tests validate the causal analysis and action selection needed for these complex tasks, with scientific and everyday applications broadening the evaluation.

Our contributions can be summarized as follows:
\begin{itemize}
  \item We introduce \sys, an LLM agent that accumulates causal knowledge from experience to develop its \cog. For each task, it combines relevant knowledge from \cog with current evidence to form \pad, which guides decisions. New observations refine both the belief and the knowledge retained for future tasks.
  \item We show that causal understanding learned from experience improves performance on complex tasks across domains. Accumulated knowledge guides what to investigate next and how to interpret new evidence, allowing earlier discoveries to improve decisions as circumstances change.
\end{itemize}
\section{Related Work}
\label{sec:related}

\paragraph{LLM Agent Systems.}

LLM agents combine reasoning, memory, planning, and tool use~\citep{wang2024survey}. ReAct~\citep{yao2023react} and Toolformer~\citep{schick2023toolformer} connect language models with external actions. ARise~\citep{zhang2025arise} combines retrieval with risk-aware search to organize reasoning, while MaR~\citep{chen2026mar} trains models to identify relevant knowledge and plan, monitor, and adjust their reasoning. System design shapes these capabilities~\citep{liu2026claudecode,openai2026harness,ning2026code}, and evaluations distinguish its contribution from that of the base model~\citep{stroebl2025hal,zhang2026harness}. MemGPT~\citep{packer2023memgpt}, retrieval-augmented generation (RAG)~\citep{lewis2020rag}, and Markdown notes in Claude Code~\citep{anthropic2026claudememory} support information retention and access beyond the current context. These mechanisms leave open how agents can, like humans, develop causal understanding from experience and use it to guide new decisions without recalling every event that shaped it.

\paragraph{Constructing Causal Knowledge.}

For causal discovery and reasoning, \citet{kiciman2024causal} examine LLM knowledge and emphasize checks against evidence. CaLM~\citep{chen2024calm} systematically evaluates these capabilities across causal tasks, prompting methods, metrics, and error types. Causal-LLM~\citep{roy2025causal}, CauScientist~\citep{peng2026causcientist}, and CauTion~\citep{peng2026caution} develop LLM-assisted methods for causal graph discovery. CauScale~\citep{peng2026causcale} scales neural causal discovery to large graphs, while CauSight~\citep{zhang2026causight} infers causal relationships among entities in images. Causal-Copilot~\citep{wang2025CausalCopilot} and the Causal AI Scientist~\citep{verma2025cais} automate causal analysis workflows, while Causal Agent~\citep{han2026causalagent} combines causal tools with iterative reasoning and graph memory. InferenceEvolve~\citep{wang2026inferenceevolve} develops and evaluates causal effect estimators. These discovery and analysis methods focus on causal structures, effect estimates, and analysis procedures. Reusing findings beyond the original analysis also requires retaining what relationship was learned and under which conditions it holds. \sys focuses on how these findings become causal understanding that can guide subsequent tasks.

\paragraph{Using and Revising Causal Knowledge.}

Causal knowledge also guides agent behavior. ADAM~\citep{yu2025adam} learns and verifies causal relationships through interaction and intervention in Minecraft, accumulating a causal graph that supports subsequent planning and task execution. Causal-aware LLMs~\citep{chen2025causalaware} learn and update environmental causal models for action selection, while CausalPlan~\citep{nguyen2025causalplan} learns causal relationships for planning and coordination. Alongside the MORE benchmark, \citet{chen2024more} introduce CAVE, an agent framework that uses causal analysis to guide multimodal reasoning and mitigate unimodal biases. CausalEvolve~\citep{chen2026causalevolve} uses a ``causal scratchpad'' to retain findings that guide program revisions, while CausalFlow~\citep{bonagiri2026causalflow} uses counterfactual analysis to repair execution failures. CauAudit~\citep{wang2026cauaudit} uses interventions to assess whether observations returned by visual tools causally affect the final answer. Together, these studies connect causal reasoning with planning, program improvement, execution repair, and the use of multimodal evidence. Yet developing lasting abstract causal cognition inspired by human learning from experience remains underexplored.
\section{Methods}
\label{sec:methods}

\subsection{Overview}
\label{sec:overview}

\begin{figure}[t]
  \centering
  \includegraphics[width=0.8\linewidth]{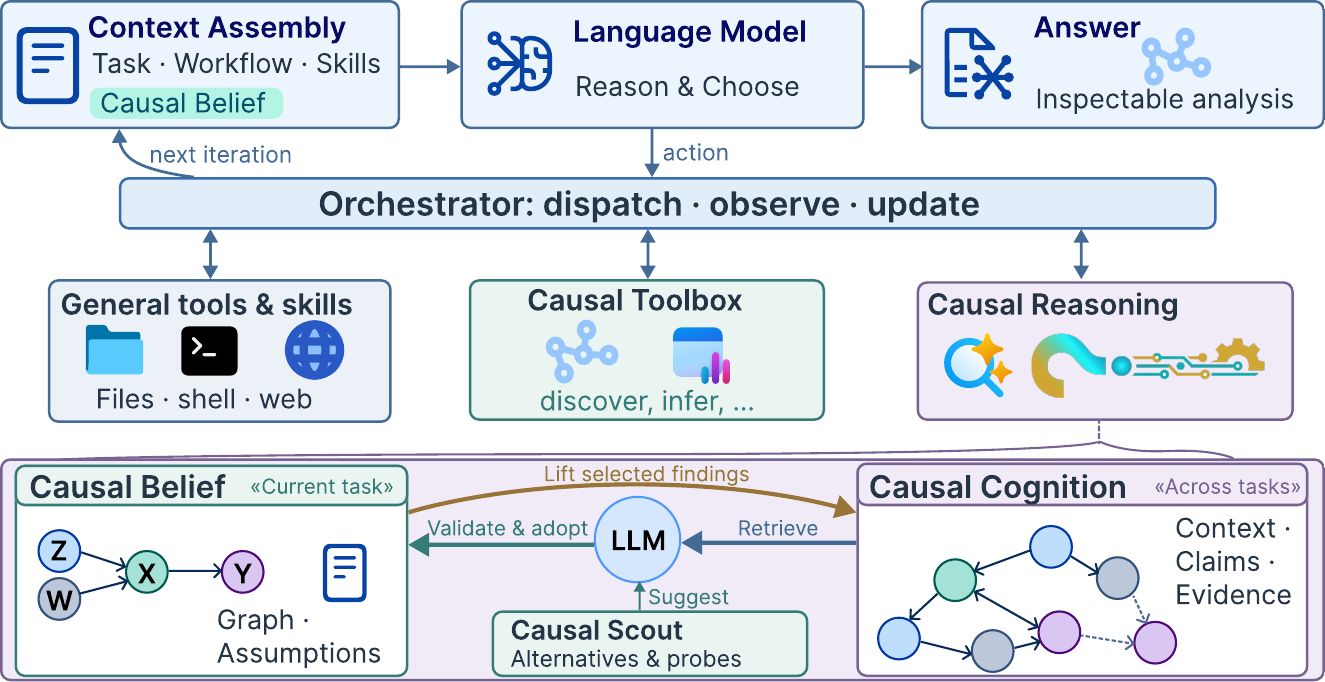}
  \caption{\textbf{Architecture of \sys.} \cog retains abstract causal knowledge across tasks, relevant parts of which are combined with the current task's evidence to form \pad.}
  \label{fig:overview}
\end{figure}

\sys learns causal relationships from experience and leverages them to guide new decisions (Figure~\ref{fig:overview}). The agent accumulates these relationships and their conditions to build \cog, reflecting how human beings develop lasting understanding from experience. To apply this accumulated understanding to a new task, the agent combines relevant knowledge from \cog with current evidence to form a structured \pad. This belief describes what influences the outcome and how possible actions could change it, guiding the agent's decisions. 

We view each task as a partially observable Markov decision process (POMDP), where the agent relies on a belief about an environment it cannot fully observe~\citep{kaelbling1998planning}. To investigate competing explanations, the agent uses \scout to compare alternatives and suggest observations that could distinguish them. As observations arrive, \sys revises the belief and updates \cog, retaining causal knowledge that remains useful beyond the individual experiences from which it was learned.

\subsection{Causal Understanding across Experiences}
\label{sec:understanding}

\paragraph{Knowledge retained across tasks.}
\cog organizes retained knowledge around explanations of how recurring factors influence one another, together with their conditions, time scales, evidence, and unresolved questions. Its graph connects these factors through proposed causal relationships. Written explanations describe mechanisms and their evidence in natural language and can be revised without first constructing a graph. In both forms, the explanation can be checked against a new task and revised as evidence changes, even after the original interaction is no longer available.

\paragraph{Applying knowledge to a task.}
When prior knowledge could help, the agent searches \cog using the user's goal and current circumstances. It follows graph connections to find relevant factors and possible paths of influence, and retrieves written explanations of related mechanisms.

The model adapts this knowledge to the task: it maps recurring factors to current variables, checks conditions and time scales against current evidence, and examines observations that challenge the explanation. Adopted relationships form the working graph in \pad; the agent also records current facts, assumptions, and open questions. For retrieved written explanations, the model records how they apply to the current situation in the belief. Thus, the same retained knowledge can inform different beliefs as task conditions and evidence change. The resulting belief guides actions, and causal analysis tools can use its graph to estimate effects.

\paragraph{Accumulating evidence.}
A retained graph relationship may receive support in one task and conflicting evidence in another. The model records these judgments as claims, assigning confidence from the available evidence.

For a relationship $r$ under specified conditions and time scale, let $Z_r\in\{0,1\}$ indicate whether it holds, with $Z_r=1$ denoting that it does. Let $Y_i$ denote a recorded claim. Assuming the claims are independent given $Z_r$, Bayesian updating gives
\begin{equation}
\label{eq:cognition-bayes}
\begin{aligned}
\operatorname{logit}(p_r) = \operatorname{logit}(\pi_r) + \sum_i\log\frac{P(Y_i\mid Z_r=1)}{P(Y_i\mid Z_r=0)}.
\end{aligned}
\end{equation}
Here $\pi_r$ and $p_r$ are the prior and updated probabilities of the relationship holding, and $\operatorname{logit}(p)=\log[p/(1-p)]$. Each likelihood ratio is a \emph{Bayes factor}, comparing how likely a claim is when the relationship holds and when it does not.

The implementation maps each claim's confidence level to a fixed weight $w_i>0$, approximating the magnitude of its log Bayes factor. Supporting claims contribute $+w_i$, and contradicting claims contribute $-w_i$:
\begin{equation}
\label{eq:cognition-update}
\begin{aligned}
L_r&=h_r+\sum_{i\in S_r}w_i-\sum_{i\in C_r}w_i, \quad
p_r&=\frac{1}{1+e^{-L_r}}.
\end{aligned}
\end{equation}
$S_r$ and $C_r$ index supporting and contradicting claims, respectively. The prior log-odds $h_r$ comes from the strongest retained hypothesis, or a weak default when none is available. Repeated proposals therefore do not accumulate as independent support. Graph retrieval follows relationships with $L_r>0$ and ranks them by $p_r$ and relevance to current conditions. Appendix~\ref{app:engineering} specifies the confidence weights and prior values.

\subsection{Revising Causal Understanding with Evidence}
\label{sec:revision}

\paragraph{Alternative explanations.}
When several explanations fit the current belief, the model proposes alternatives for \scout to compare. Individually plausible relationships may become incompatible when combined.

\scout uses Answer Set Programming (ASP)~\citep{gebser2019clingo} to enumerate combinations. ASP specifies choices and constraints; each solution, called an \emph{answer set}, describes a candidate graph satisfying them. Rules enforce required and forbidden relationships, exclude incompatible hypotheses, prevent directed cycles, and respect known temporal order.

Candidates are also checked against available statistical test results. \scout ranks alternatives by confidence and simplicity and returns supporting evidence, counterevidence, and observations that could distinguish them. The model uses these suggestions to decide what to investigate and which relationships to adopt.

\paragraph{Revising the current belief.}
As new evidence arrives, the model compares it with its explanation and expected outcomes, keeping observations distinct from predictions and checking their conditions and time periods. It can add or remove relationships, change assumptions, or revise the conditions under which an explanation applies. These revisions inform subsequent action selection and investigation.

With suitable data, causal analysis tools help discover relationships, select factors to account for when estimating effects, estimate those effects, and check their reliability. Results return to \pad, allowing the model to revisit its explanation or choose further analyses.

\paragraph{Retaining revised knowledge.}
The agent selects findings that may help later tasks and expresses them in terms of recurring factors, preserving the conditions and evidence that make the explanation applicable. For graph knowledge, \emph{Lift} maps selected task variables back to factors in \cog and records the relationship, its mechanism, conditions, and supporting or conflicting evidence.

Findings can remain hypotheses. Claims about an existing relationship under the same conditions and time scale update its evidence and score through Equation~\ref{eq:cognition-update}. Findings under different conditions or time scales are retained separately. These changes affect which explanations are retrieved for later tasks.

Written explanations are revised directly to reflect changed mechanisms, conditions, or uncertainty. Earlier versions and reasons for change are retained; retrieval selects the latest version available at the task's time.

In a nutshell, \sys applies retained understanding to a task, revises it against current evidence, and preserves selected changes for later decisions. The agent chooses when to read, retain, and revise this knowledge.
Selected system and tool instructions are provided in Appendix~\ref{app:prompts}.
\section{Experiments}
\label{sec:experiments}

\subsection{Settings}
\label{sec:exp-setup}

Unless otherwise specified, we use DeepSeek-V4-Pro~\citep{deepseek2026v4} in all experiments because it is open-source, supports reproducibility, and offers strong overall performance at a low cost. Our baselines are ReAct~\citep{yao2023react} and OpenCode~\citep{opencode2026github}, an open-source harness with over 200,000 GitHub stars. Financial studies additionally include text-memory ReAct; the tabular study includes model-only references. Chronological financial tasks start in fresh contexts with shared candidate and model-call limits. Appendix~\ref{app:experimental-details} provides supplementary settings, metric definitions, and detailed results.

\subsection{Developing Causal Cognition through Experience}
\label{sec:exp-development}

\begin{figure}[t]
  \centering
  \includegraphics[width=\textwidth]{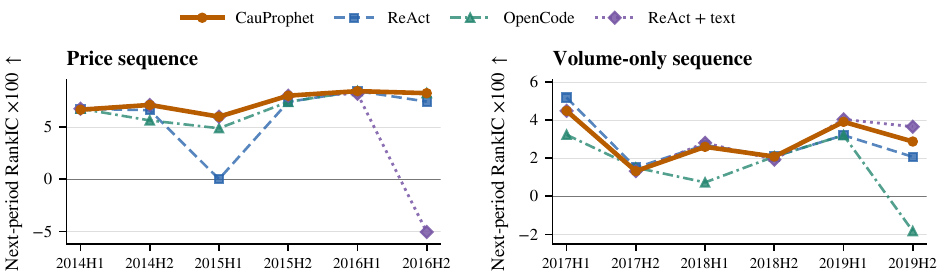}
  \caption{\textbf{Financial research across successive periods.} \sys achieves the highest mean score across the two studies. Each point scores a committed formula on the next half-year.}
  \label{fig:exp-longitudinal}
\end{figure}

To study how agents develop and reuse causal knowledge over time, we use AlphaBench~\citep{luo2026alphabench}. Its factor-search task asks agents to construct formulas that rank stocks by their predicted next-day returns, combining numerical trends with hypotheses about market behavior. We follow two six-period studies---price-and-volume research in 2014--2016 and volume-only research in 2017--2019---in which the agent commits a formula before receiving next-period feedback (Figure~\ref{fig:exp-longitudinal}). Each task starts in a fresh context; agents with memory can retrieve earlier findings and revise them after feedback.

\begin{table}[t]
  \centering
  \small
  \setlength{\tabcolsep}{4pt}
  \begin{tabularx}{0.65\linewidth}{>{\raggedright\arraybackslash}Xrrr}
    \toprule
    & \multicolumn{3}{c}{Next-period RankIC $\times 100$ $\uparrow$} \\
    \cmidrule(lr){2-4}
\textbf{Method} & \textbf{Price} & \textbf{Volume} & \textbf{Mean} \\
    \midrule
ReAct & 6.119 & 2.792 & 4.455 \\
OpenCode & 6.901 & 1.488 & 4.195 \\
ReAct + text memory & 5.172 & \bestcell{3.041} & 4.106 \\
\rowcolor{TableFocus}
\textbf{\sys} & \bestcell{7.418} & 2.881 & \bestcell{5.149} \\
    \bottomrule
  \end{tabularx}
  \caption{Chronological factor research, RankIC $\times 100$. We use 694/722 common dates for price/volume, average within each half-year, then weight periods and studies equally. Blue cells mark column maxima.}
  \label{tab:exp-alphabench}
\end{table}

RankIC measures agreement between predicted and realized return rankings. Across the two chronological studies, \sys achieves the highest mean score, outperforming ReAct, OpenCode, and ReAct with text memory (Table~\ref{tab:exp-alphabench}). In price research, \sys retains its hypothesis that a five-day decline may precede a rebound after testing alternatives. In volume research, \sys initially retains the hypothesis that stocks with trading volume below their historical average may have higher next-day returns. Later feedback shows that this hypothesis performs differently across periods. The agent incorporates this finding into its retained explanation. The updated explanation then guides tests of how to combine recent and longer-term trading volume and informs revisions to the prediction formula.

\begin{figure}[t]
  \centering
  \includegraphics[width=0.66\textwidth]{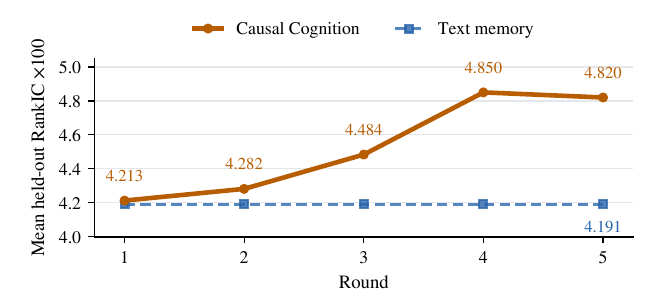}
  \caption{\textbf{Knowledge reuse across five rounds of formula search.} Mean held-out RankIC $\times 100$ for \cog and text memory; higher is better.}
  \label{fig:exp-alphabench-rounds}
\end{figure}

A complementary five-round study examines knowledge reuse during repeated formula search on the same historical dataset (Figure~\ref{fig:exp-alphabench-rounds}). The \cog workflow's mean held-out score rises from 4.213 to 4.850 in Round~4 and finishes at 4.820, while text memory remains at 4.191 across all five rounds. Text memory selects the same final formula in every round, whereas the \cog workflow uses retained comparisons to continue testing and revising formulas. Full scores and search effort are in Tables~\ref{tab:alphabench-memory} and~\ref{tab:alphabench-starts}.

\begin{figure}[t]
  \centering
  \includegraphics[width=0.72\textwidth]{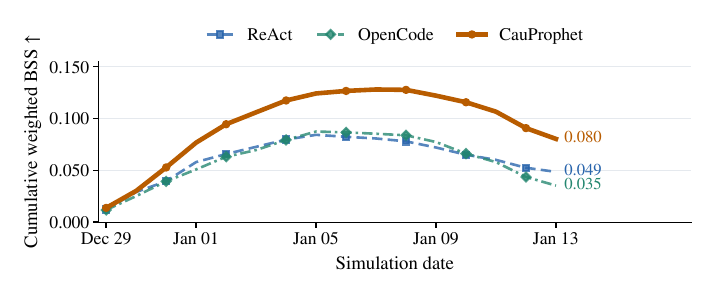}
  \caption{\textbf{FutureSim over time.} \sys achieves the highest overall BSS before events resolve. Curves show cumulative question-weighted BSS for pre-resolution predictions; endpoints match Table~\ref{tab:exp-futuresim}.}
  \label{fig:exp-futuresim-curve}
\end{figure}

We further examine whether causal understanding helps agents draw inferences across diverse domains as evidence accumulates over time. Using FutureSim~\citep{goel2026futuresim,openforecaster2026futuresim}, we follow 24 OpenForesight forecasting questions~\citep{chandak2026openforesight} as news becomes available in chronological order. This yields 195 question--day predictions per system, including 171 before resolution. Brier Skill Score (BSS) rewards probability assigned to the eventual answer and penalizes confident mistakes; pre-resolution scores average within each question, then across questions. Figure~\ref{fig:exp-futuresim-curve} tracks prediction quality as the replay unfolds. \sys achieves the highest BSS and Top-1 accuracy before events resolve (Table~\ref{tab:exp-futuresim}). This advantage concerns judgments made while an event is unfolding, when decisions still need to be made.

\begin{table}[t]
  \centering
  \small
  \setlength{\tabcolsep}{4pt}
  \begin{tabularx}{\linewidth}{>{\raggedright\arraybackslash}Xrrrrrr}
    \toprule
    & \multicolumn{3}{c}{Before resolution} & \multicolumn{3}{c}{Final snapshot} \\
    \cmidrule(lr){2-4}\cmidrule(lr){5-7}
\textbf{Method} & \textbf{BSS $\uparrow$} & \textbf{Top-1 (\%) $\uparrow$} & \textbf{Truth prob. (\%) $\uparrow$} & \textbf{BSS $\uparrow$} & \textbf{Top-1 (\%) $\uparrow$} & \textbf{Truth prob. (\%) $\uparrow$} \\
    \midrule
ReAct & 0.049 & 21.3 & 13.2 & \bestcell{0.092} & 29.2 & \bestcell{16.9} \\
OpenCode & 0.035 & 22.0 & 12.6 & 0.006 & 29.2 & 15.0 \\
\rowcolor{TableFocus}
\textbf{\sys} & \bestcell{0.080} & \bestcell{28.6} & \bestcell{14.7} & 0.085 & 29.2 & 16.3 \\
    \bottomrule
  \end{tabularx}
  \caption{FutureSim over 24 questions. Top-1 measures whether the highest-probability answer is correct; truth probability is the probability assigned to the resolved answer. Final scores use resolution-day snapshots. Blue cells mark column maxima where methods differ.}
  \label{tab:exp-futuresim}
\end{table}

\begin{figure}[t]
  \centering
  \includegraphics[width=\textwidth]{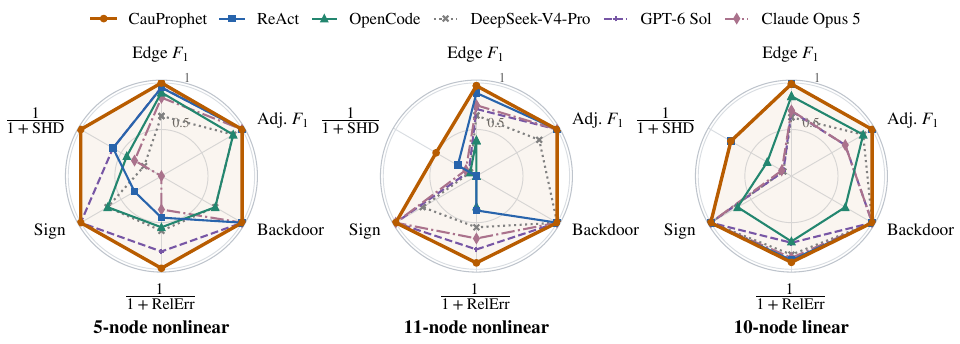}
  \caption{\textbf{Causal analysis on three SCMs.} \sys achieves the lowest mean relative error in effect estimation across all three systems. Outward is better: $F_1$ scores and Backdoor/Sign success rates use $[0,1]$; SHD and fractional relative error $e$ use $1/(1+e)$. Curves show transformed means; raw means and SDs are in Table~\ref{tab:rq1_compcheck}.}
  \label{fig:exp-scm}
\end{figure}

\subsection{Foundational Causal Abilities}
\label{sec:exp-causal}

Causal reasoning in the complex tasks above depends on basic abilities to identify causal relationships, estimate how actions affect outcomes, and choose interventions. We therefore test these foundations directly using fixed datasets with known causal mechanisms and an interactive environment.

To test causal analysis against known answers, we use data from three structural causal models (SCMs), whose equations define the causal relationships and intervention effects. They include two nonlinear systems and one linear system, each with 800 observations. \sys has the lowest mean effect relative error in all three and gets the effect direction correct in all nine evaluations (Figure~\ref{fig:exp-scm}). ReAct finds a valid set of variables to account for when estimating effects in the eleven-node system but reverses every effect direction. Accurate causal analysis requires both recovering relationships and estimating how interventions change the outcome.

\begin{figure}[t]
  \centering
  \includegraphics[width=0.85\textwidth]{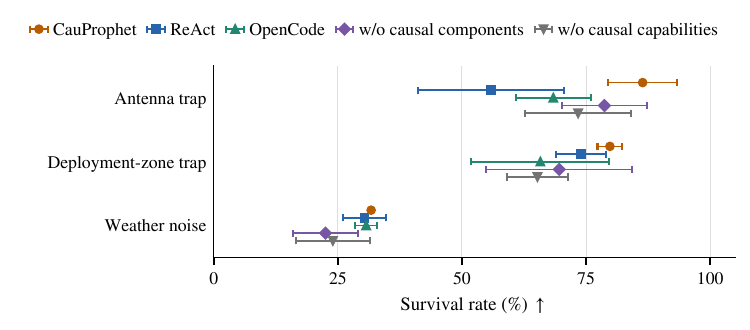}
  \caption{\textbf{CausalGame: survival in a simulated drone adventure.} \sys achieves the highest mean survival across all three scenarios. Means and sample SDs over three runs per scenario and configuration. Scores and victories are retained in Table~\ref{tab:exp-causalgame}.}
  \label{fig:exp-causalgame}
\end{figure}

We then move from causal structure discovery and effect estimation on fixed datasets to choosing interventions in an interactive environment. We use CausalGame~\citep{chen2026causalgame}, a simulated drone adventure, to test whether causal understanding guides actions when agents have only partial information about their surroundings. \sys has higher mean survival and score than ReAct and OpenCode in all three scenarios, winning every run in both trap scenarios (Figure~\ref{fig:exp-causalgame}). The complete system also exceeds both ablations: joint removal of \pad, \cog, and \scout with causal computation retained, and removal of all dedicated causal capabilities. The results connect causal analysis to the consequences of actions; computation tools alone do not recover the complete system's performance.

\begin{table}[t]
  \centering
  \small
  \setlength{\tabcolsep}{5pt}
  \begin{tabularx}{0.92\linewidth}{>{\raggedright\arraybackslash}Xrrrr}
    \toprule
    \textbf{Method} & \textbf{Model} & \textbf{Variables} & \textbf{Value transform} & \textbf{Graph transform} \\
    \midrule
    \textbf{\sys} & \bestcell{91.7} & \secondcell{87.7} & \bestcell{48.5} & \bestcell{61.6} \\
    ReAct & \bestcell{91.7} & \thirdcell{82.9} & \secondcell{40.3} & \thirdcell{46.5} \\
    OpenCode & \secondcell{66.7} & \bestcell{88.7} & \thirdcell{37.2} & \secondcell{58.5} \\
    \bottomrule
  \end{tabularx}
  \caption{\textbf{Scientific analysis on 12 BLADE tasks.} Mean reference-match rates (\%; higher is better). Transformation graphs describe data processing. Blue shades, from darker to lighter, mark the three highest distinct scores in each column; best scores are bold and ties share a shade.}
  \label{tab:exp-blade}
\end{table}

\begin{table}[t]
  \centering
  \small
  \setlength{\tabcolsep}{3pt}
  \begin{tabularx}{0.65\linewidth}{>{\raggedright\arraybackslash}Xrrr}
    \toprule
\textbf{Task category} & \textbf{Tasks} & \textbf{Executions} & \textbf{Checks} \\
    \midrule
Music & 5 & 9 & 74/74 \\
Notes and messaging & 1 & 1 & 8/8 \\
Payments and expenses & 4 & 6 & 29/29 \\
Files and photos & 3 & 4 & 32/32 \\
Shopping & 3 & 3 & 30/31 \\
    \midrule
\rowcolor{TableFocus}
\textbf{Total} & \textbf{16} & \textbf{23} & \textbf{173/174} \\
    \bottomrule
  \end{tabularx}
  \caption{AppWorld results for \sys on selected tasks. Checks are passed/total checks across recorded executions; per-task outcomes are in Table~\ref{tab:appworld_tasks}.}
  \label{tab:exp-appworld}
\end{table}

\subsection{Causal Reasoning across Diverse Domains}
\label{sec:exp-domains}

To examine how agents answer scientific questions through data analysis, we evaluate 12 tasks from BLADE~\citep{gu2024blade}, spanning education, social behavior, ecology, health, and economic decisions. Model match measures agreement with the reference statistical model; variable match measures agreement on the selected variables. Transformation match checks the resulting data values and how the processing steps connect. \sys matches the best model selection, achieves the highest transformation match rates, and improves variable match over ReAct (Table~\ref{tab:exp-blade}). The clearest gains concern data preparation: causal questions must be translated into the observations and comparisons that answer them.

\Needspace{6\baselineskip}
Finally, we examine action selection in everyday applications using AppWorld~\citep{trivedi2024appworld}, with tasks involving music, messaging, expenses, files, and shopping. Across 23 completed executions of 16 selected tasks, \sys passes 173 of 174 checks (Table~\ref{tab:exp-appworld}). These tasks connect action selection to changes in everyday applications.

Appendix~\ref{app:interactive-case} opens with four forecasts recorded before their outcomes became public, then shows how causal reasoning guides a shopper's decision and helps interpret a classroom experiment.

\FloatBarrier
\section{Conclusion}

We introduced \sys to study how agents can build causal understanding from experience, much as human beings do. By connecting \cog with \pad, the agent applies retained causal relationships and their conditions to new situations, combining them with current evidence to guide decisions. Findings from these tasks can contribute to the causal knowledge available for later use. Experiments show that earlier findings guide subsequent research and that \sys makes more accurate predictions while events unfold. Further evaluations examine the basic causal abilities underlying these tasks and applications to scientific analysis and everyday activities. These results support our approach to developing agents whose causal understanding extends beyond individual experiences and informs new decisions.
\section*{Limitations}
\phantomsection
\addcontentsline{toc}{section}{Limitations}
When evidence is insufficient to test causal relationships, \sys relies more heavily on the base model's prior knowledge. Richer observations and feedback from actions could help assess the reliability of the knowledge it accumulates. Our experiments cover a limited number of stages; longer studies in changing real-world settings could examine how causal understanding continues to grow and adapt over time.

\FloatBarrier



\bibliography{custom}
\clearpage
\appendix
\section{Case Studies}
\label{app:interactive-case}

The figures in this section show how \sys uses causal reasoning across a range of real-world applications. We begin with four forecasts recorded before the outcomes were known. We then follow a shopper deciding whether to buy now or wait and examine whether students who practice with AI perform better on their own. Each case began with a single user request to \sys running Claude Opus~5.5.

\subsection{Forecasting Real-World Events with Causal Reasoning}
\label{app:forecast-cases}

Before an election is called or a price index is released, which of today's signals will matter? \sys traces how those signals could shape the outcome, weighs competing influences, and commits to a probability while the outcome is still unknown. Readers can then check both the prediction and, when the release gives a breakdown, the reasoning behind it.

Each figure pairs the original forecast with the outcome we verified after publication.

\begin{figure}[p]
  \centering
  \includegraphics[width=\textwidth,height=0.87\textheight,keepaspectratio]{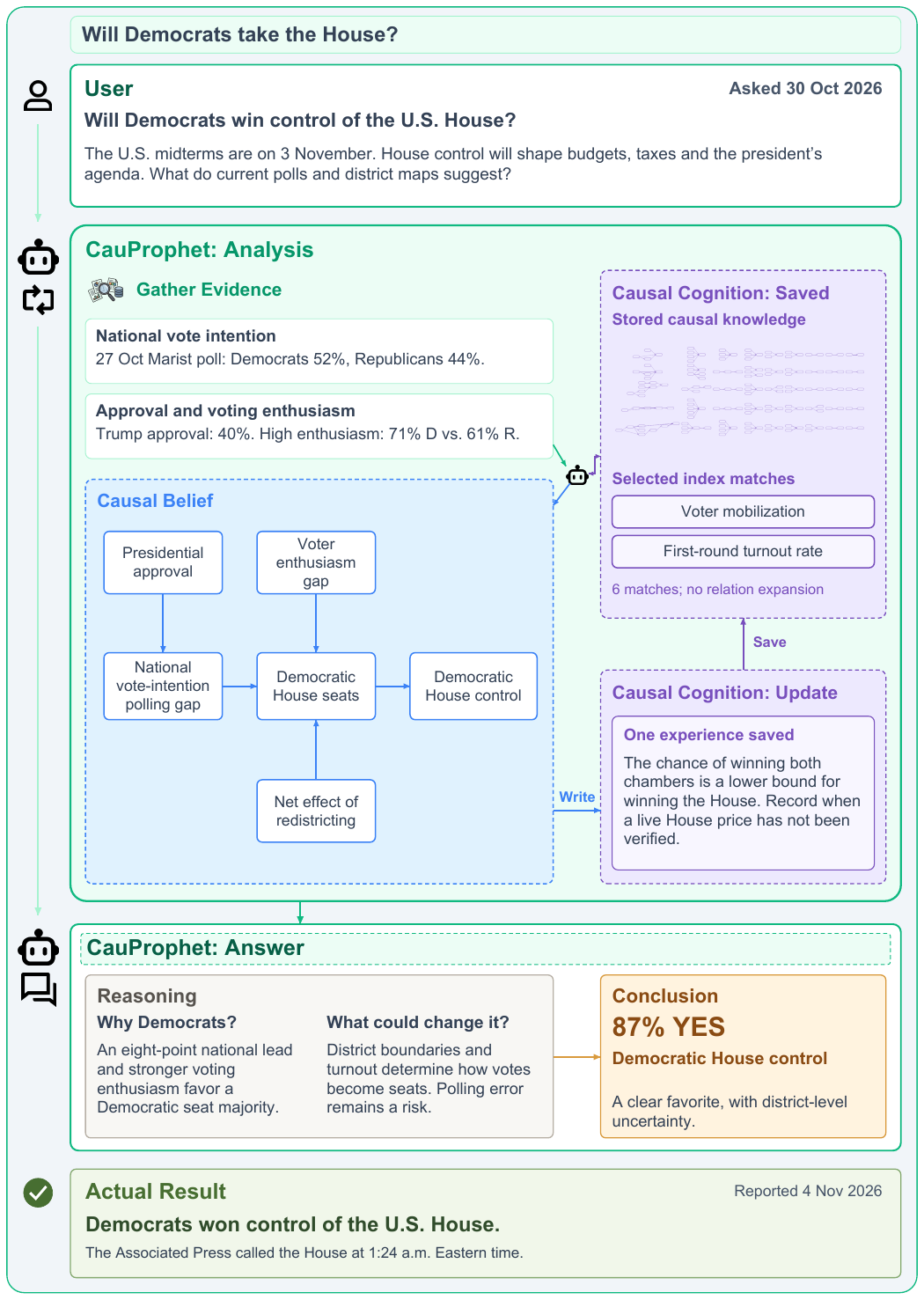}
  \caption{\textbf{Who would win the U.S. House?} The October 30 forecast gave Democrats an 87\% chance of winning control. The Associated Press called the House for Democrats on November~4.}
  \label{fig:forecast-us-house}
\end{figure}

\begin{figure}[p]
  \centering
  \includegraphics[width=\textwidth,height=0.87\textheight,keepaspectratio]{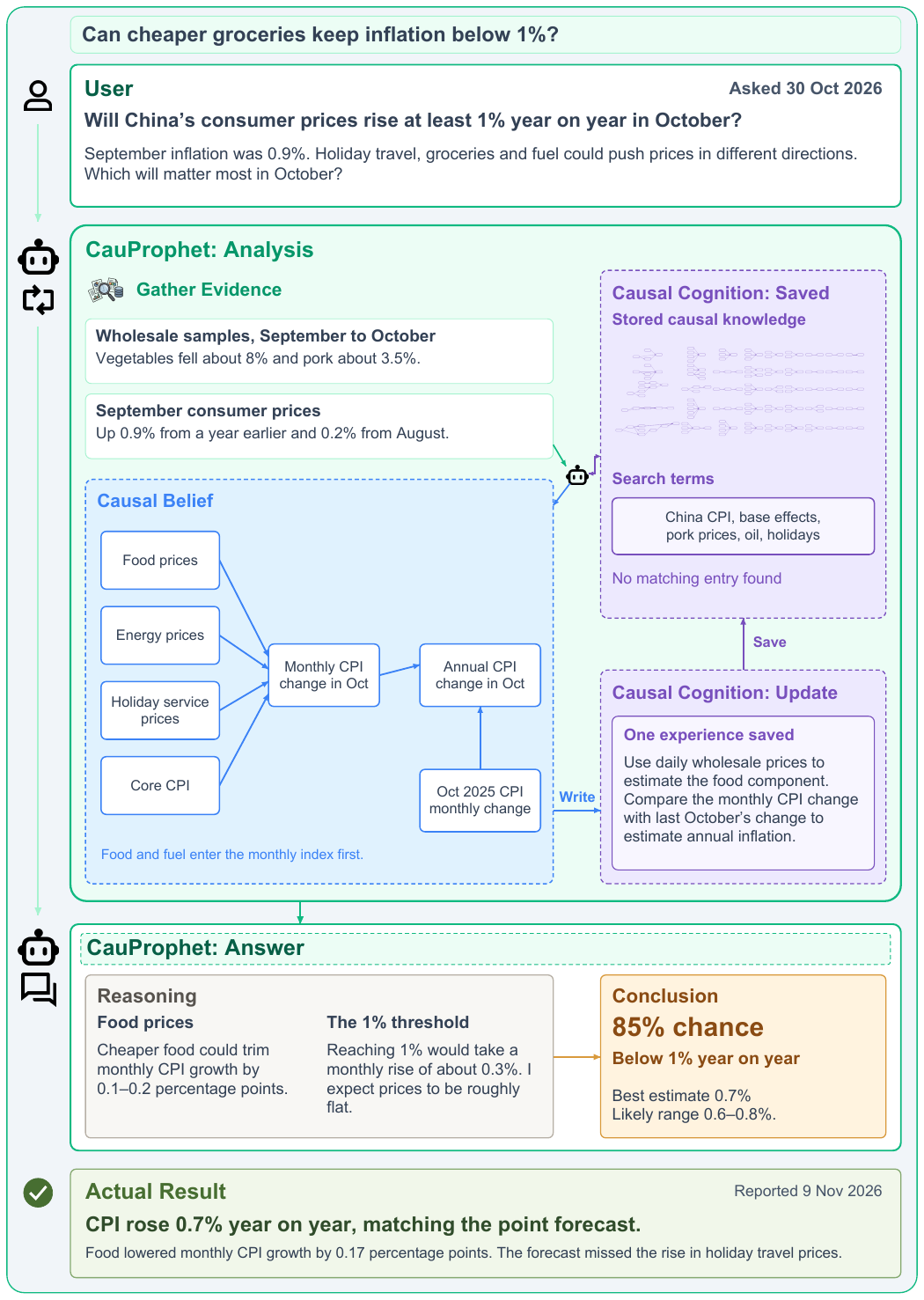}
  \caption{\textbf{Would consumer price inflation in China reach 1\%?} The October 30 forecast put annual inflation at 0.7\%. China's National Bureau of Statistics reported that value on November~9. Food prices reduced monthly inflation by about 0.17 percentage points, within the forecast range of 0.1--0.2 points.}
  \label{fig:forecast-china-cpi}
\end{figure}

\begin{figure}[p]
  \centering
  \includegraphics[width=\textwidth,height=0.87\textheight,keepaspectratio]{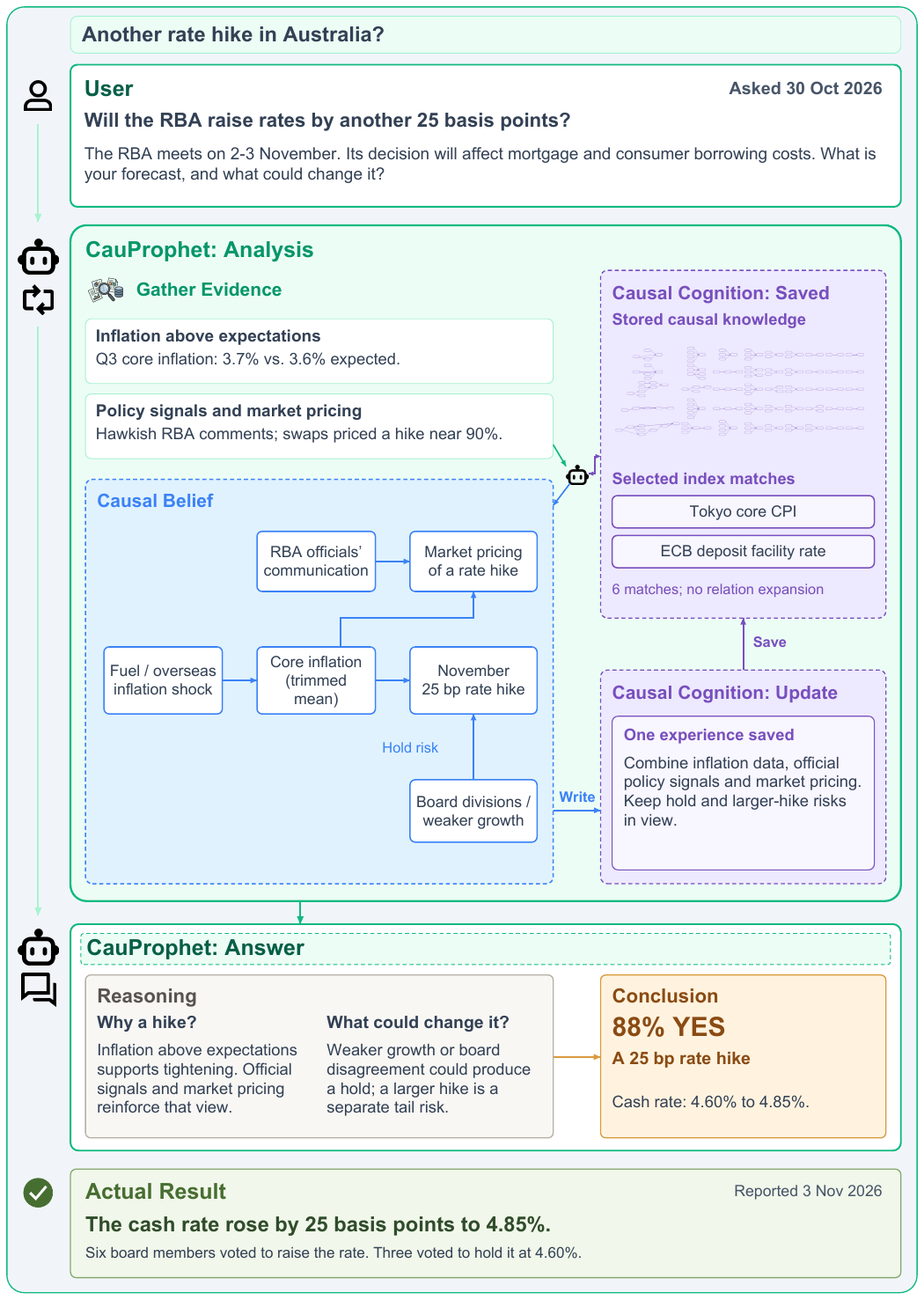}
  \caption{\textbf{Would Australia raise interest rates again?} The Reserve Bank of Australia voted six to three on November~3 to raise the cash rate by 25 basis points to 4.85\%, the outcome to which the October 30 forecast assigned an 88\% probability.}
  \label{fig:forecast-rba}
\end{figure}

\begin{figure}[p]
  \centering
  \includegraphics[width=\textwidth,height=0.87\textheight,keepaspectratio]{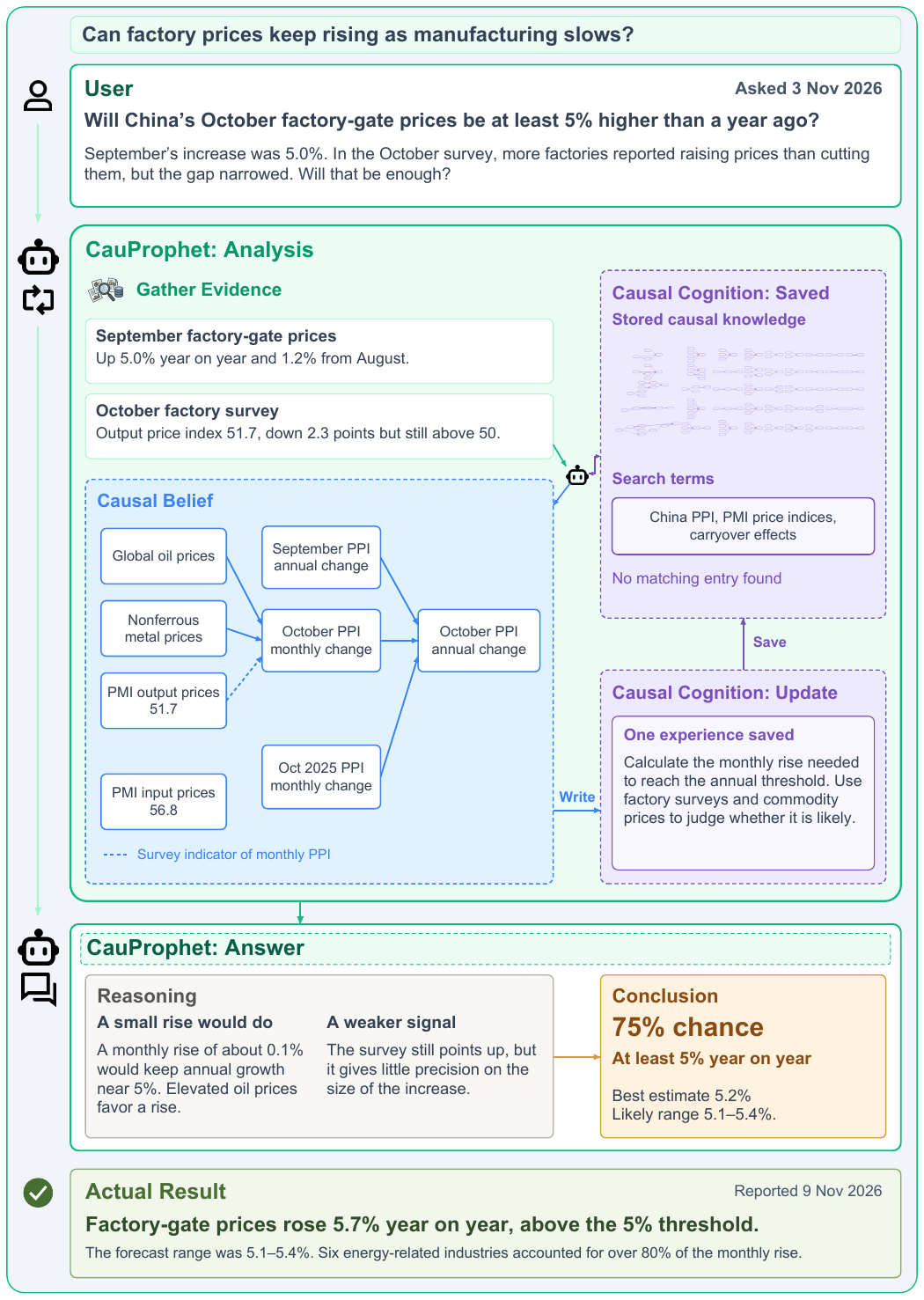}
  \caption{\textbf{Would Chinese factory-gate prices rise at least 5\% year on year?} The November 3 forecast assigned a 75\% probability to crossing or meeting the threshold. The National Bureau of Statistics reported a 5.7\% rise on November~9, above the estimated range of 5.1--5.4\%.}
  \label{fig:forecast-china-ppi}
\end{figure}

For the U.S. House, \sys weighs the Democrats' national polling lead against the way district boundaries translate votes into seats (Figure~\ref{fig:forecast-us-house}). For China's consumer prices, it uses wholesale food prices to estimate how much cheaper groceries will pull down the monthly index (Figure~\ref{fig:forecast-china-cpi}). For Australian interest rates, it weighs stronger-than-expected underlying inflation against the possibility that a divided board will hold rates steady (Figure~\ref{fig:forecast-rba}). For Chinese producer prices, it uses a manufacturing survey to judge whether factories are still raising prices (Figure~\ref{fig:forecast-china-ppi}).

The House and interest-rate searches returned index matches in \cog. Neither price-index search found a matching entry. In each run, the agent formed a \pad from the available evidence and saved one written experience when making its forecast.

\FloatBarrier

\subsection{Guiding Everyday Decisions Under Uncertainty}
\label{app:case-labubu}

Should I act now, or is it worth waiting? The same observation can point to different choices depending on what caused it. \sys weighs those explanations against the evidence and identifies what a user can check before deciding. The recommendation takes the user's budget, preferences, and deadline into account.

\begin{figure}[tp]
  \centering
  \includegraphics[width=\textwidth,height=0.87\textheight,keepaspectratio]{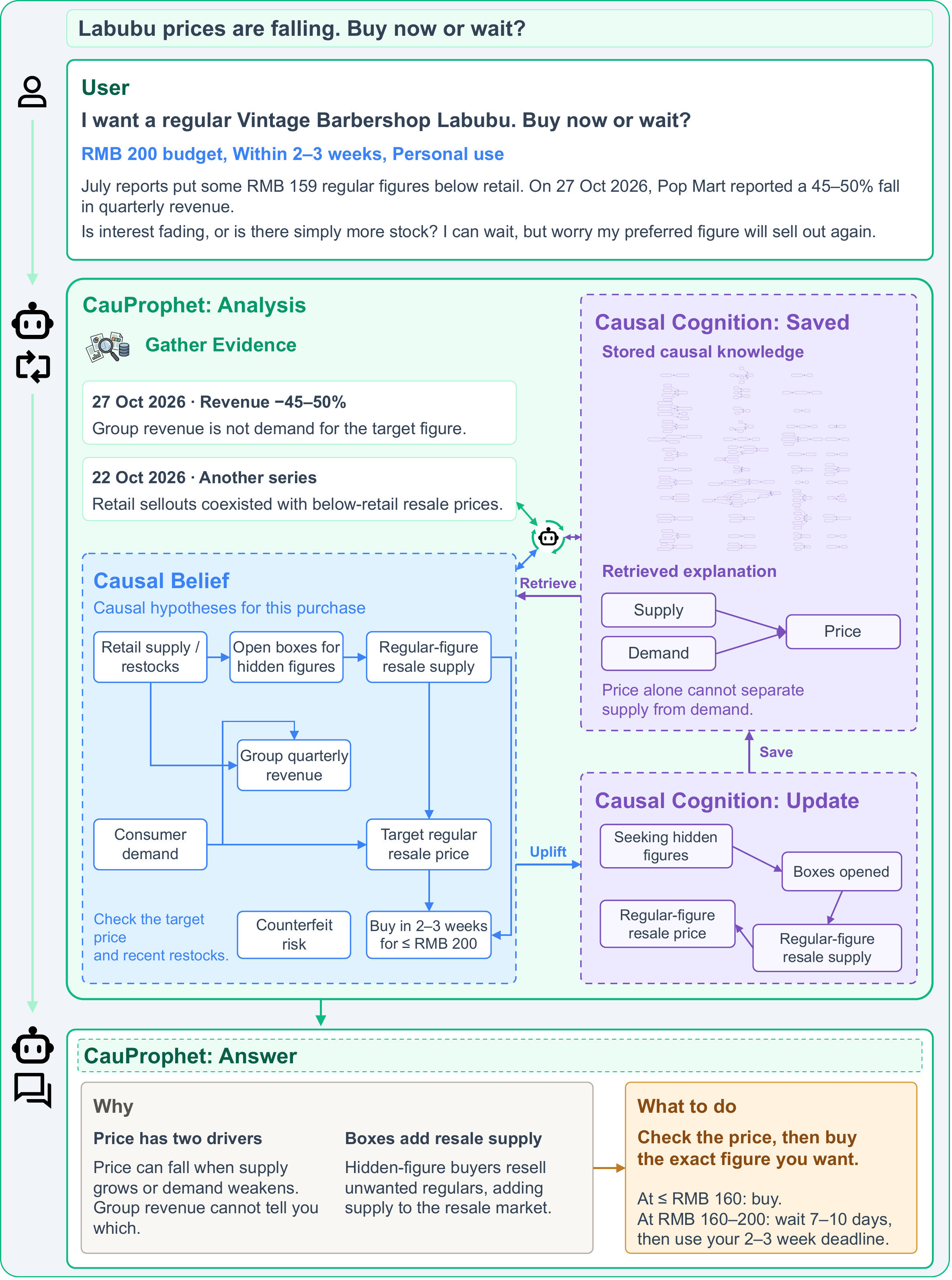}
  \caption{\textbf{Should I buy a Labubu now or wait?} Purple denotes \cog and blue denotes \pad. The faded graph provides context for stored knowledge. The purple diagrams visualize a retrieved written explanation and the newly saved text experience; the blue graph shows the complete recorded belief. Arrows connect knowledge retrieval, analysis, and experience retention within the single-request conversation.}
  \label{fig:labubu-purchase}
\end{figure}

In the October~28, 2026 conversation, a user wants a regular Labubu figure from the Vintage Barbershop series for personal use, with a budget of RMB~200 and a deadline of two to three weeks. Reports of below-retail resale prices and declining Pop Mart quarterly revenue make the user wonder whether demand is fading or supply is growing, and whether waiting could mean missing the desired figure (Figure~\ref{fig:labubu-purchase}).

\sys retrieves a written explanation from \cog: supply and demand jointly determine price, so a price decline alone cannot identify which side changed. It then searches recent reporting, distinguishing group revenue from demand for the user's particular figure and finding that initial retail sellouts can coexist with below-retail resale listings in another Labubu series. These observations motivate a proposed mechanism: buyers seeking rare hidden figures open blind boxes and resell unwanted regular figures, increasing their resale supply even when unopened boxes sell out quickly.

The recorded \pad connects retail supply, box opening, regular-figure resale supply, consumer demand, group revenue, and the target price to buying within the user's budget and deadline. All eight variables and eight directed edges appear in the figure, including the separately recorded counterfeit risk. The answer recommends checking the exact figure's price and buying it with its identity already known. It proposes buying at about RMB~160 or below; at RMB~160--200, it suggests waiting seven to ten days before deciding within the original deadline. The current transaction price and recent restocks are checks for carrying out this plan.

Before answering, \sys saves one new experience in \cog, retaining the proposed box-opening mechanism and the distinction between group revenue and the availability of a particular figure. The purple update diagram visualizes this saved text. The case shows how a general supply--demand explanation informs a concrete purchase decision and how the resulting reasoning becomes an experience available for later retrieval.

\FloatBarrier

\subsection{Testing Scientific Claims with Experimental Data}
\label{app:case-math}

What would count as evidence that an intervention worked? \sys translates a scientific question into a comparison the experiment can support. It identifies who received the intervention, when the outcome was measured, and which comparison answers the original question.

\begin{figure}[tp]
  \centering
  \includegraphics[width=\textwidth,height=0.87\textheight,keepaspectratio]{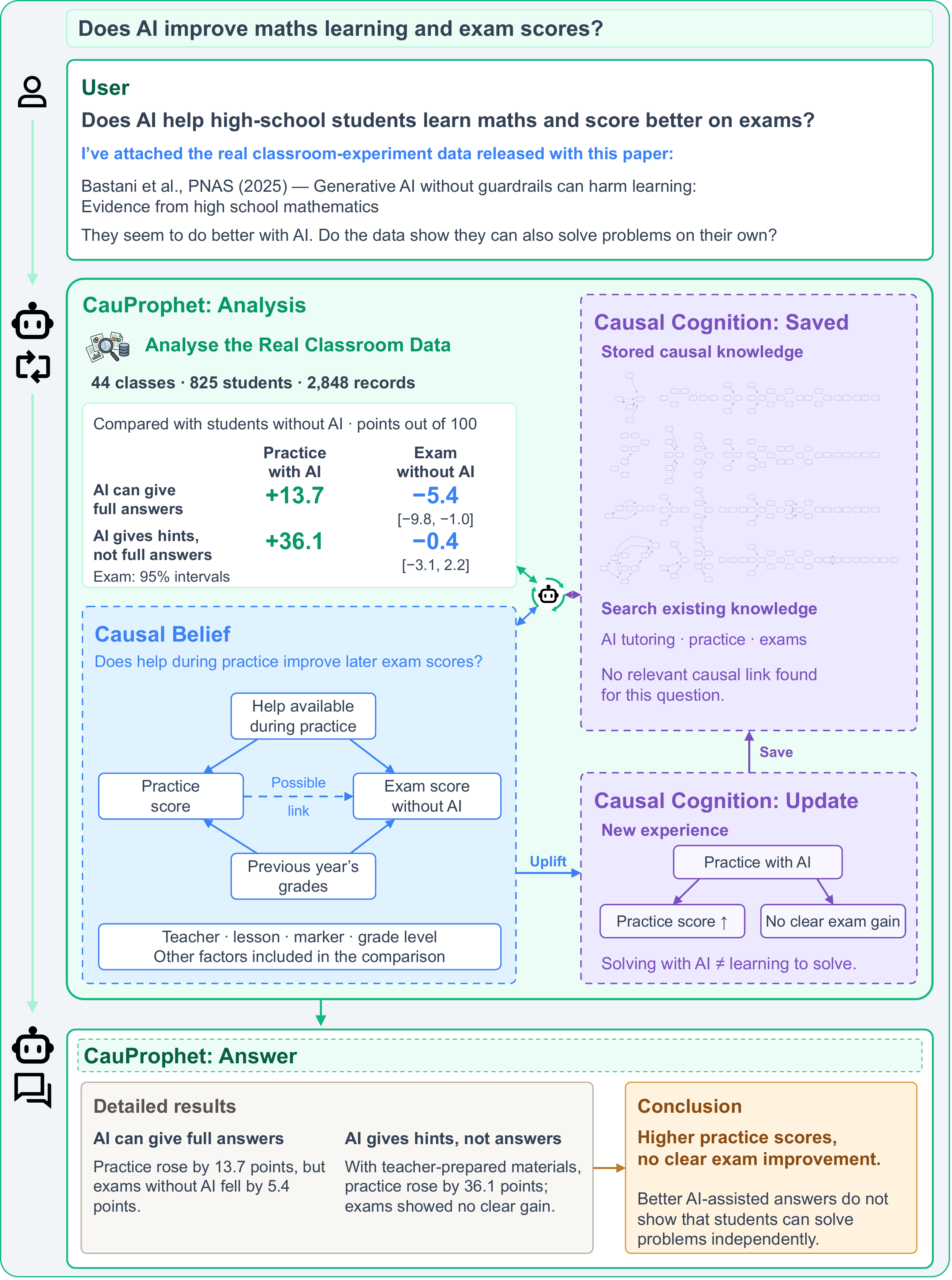}
  \caption{\textbf{Does AI improve mathematics learning and exam scores?} The user supplies real classroom data from \citet{bastani2025generative}. Values are adjusted differences from students without AI, rescaled to 100 points; brackets show 95\% intervals for exam effects. Blue shows the complete recorded \pad; the dashed edge is a hypothesized link. Purple shows stored \cog and a diagram of the newly saved text experience.}
  \label{fig:math-learning}
\end{figure}

In the October~29, 2026 conversation, the user supplies real classroom-experiment data released with \citet{bastani2025generative} and asks whether using AI helps high-school students learn mathematics and perform better on their own. The study compares practice with an AI that can give full answers, an AI designed to provide hints with teacher-prepared materials, and no AI. All groups subsequently take an exam without AI. The supplied records and study description let \sys distinguish performance with assistance from independent learning (Figure~\ref{fig:math-learning}).\footnote{Author-released data and analysis code: \url{https://github.com/obastani/GenAICanHarmLearning}.}

\Needspace{7\baselineskip}
\sys inspects the data and executes statistical analyses. Following the study's main analysis, it excludes honors classes and uses complete observations for the comparison: 2,848 student--lesson records from 825 students in 44 classes. It accounts for previous grades, teacher, lesson, marker, and grade level, with uncertainty calculated at the class level. Because practice scores are already affected by AI use, the agent does not adjust for them when estimating the effect on the subsequent exam.

The comparison yields a clear contrast. Relative to students without AI, practice scores increase by 13.7 points for the AI that can give answers and by 36.1 points for the AI that provides hints, after rescaling scores to 100 points. On the exam without AI, the corresponding differences are $-5.4$ points (95\% interval: $[-9.8,-1.0]$) and $-0.4$ points ($[-3.1,2.2]$). The answer therefore reports higher practice scores but no clear improvement in independent exam performance; the group whose AI could give full answers scores lower. Better performance while AI is available does not by itself establish that students have learned to solve the problems independently.

The agent records five variables and five directed edges in \pad. The possible link from practice to exam performance is a hypothesis, shown dashed, while the adjustment factors remain separately recorded. Its search of \cog returns no directly applicable causal relationship. After the analysis, it saves one new text experience: distinguish AI-assisted practice gains from independent exam gains, and evaluate learning using performance without AI. The purple update diagram illustrates that experience. The mathematics and purchase cases each add a written experience to \cog; the persistent graph's existing concepts and relationships retain their recorded state.

\FloatBarrier

\section{Method and Implementation Details}
\label{app:engineering}

\subsection{Representing Causal Knowledge}

An experience can suggest that one factor influences another under particular conditions. \cog retains the recurring factors and proposed relationship, together with a mechanism, time scale, and supporting or conflicting evidence. For graph knowledge, a source link connects the relationship to the experience that motivated it. In the Labubu case in Appendix~\ref{app:case-labubu}, the written explanation distinguishes supply from demand when interpreting a resale-price decline. The \pad also includes the variables and observations needed to interpret the current reports. Written explanations provide another way to retain a mechanism and its conditions, including when the agent has not constructed a graph.

\subsection{Applying Knowledge to a New Task}

When retained knowledge could help, the agent searches for relevant concepts, graph connections, or written explanations. It maps recurring factors to variables in the current situation and examines whether the retained conditions and time scale still fit. For example, applying the resale-price explanation requires checking whether production, available resale offers, or buyers' demand has changed. Relationships the model adopts enter the working graph in \pad. When using a written explanation, the model records the selected version and its own account of how it applies to the present task. These judgments guide the next reply or investigation.

\subsection{Revising Knowledge with Evidence}

New observations may support an explanation, contradict it, or reveal a missing condition. The agent revises the belief and selects updates to retained knowledge.

For the low, medium, and high confidence levels, the evidence weights $w_i$ in Equation~\ref{eq:cognition-update} are $0.5$, $1$, and $2$, respectively. Hypotheses at these levels assign prior log-odds of $0.25$, $0.75$, and $1.25$; the largest is used, with $h_r=0.25$ when no hypothesis is recorded. Claims are aggregated for a fixed directed relation, set of conditions, and time scale. Each source episode contributes at most one claim to that relation; revising the claim replaces its previous verdict and confidence.

Findings under different conditions or time scales are retained separately. Written explanations are revised directly, preserving the reason for the change and earlier versions. Retrieval selects the latest version available at the task's time.

\subsection{Comparing Alternative Explanations}

A current belief can admit several explanations of the same observation. The model proposes hypotheses for \scout to compare. Each candidate combines selected hypotheses with the known relationships. \scout checks the resulting graph against supplied constraints on relationships, temporal order, and available statistical evidence, then ranks alternatives by confidence and simplicity. Its response associates each alternative with supporting evidence, counterevidence, and observations that could distinguish it. The model uses this comparison to choose which explanation to investigate or adopt in the current belief.

\FloatBarrier
\Needspace{10\baselineskip}
\section{Experimental Details and Additional Analyses}
\label{sec:appendix}
\label{app:experimental-details}


\subsection{AlphaBench}
\subsubsection{Chronological Studies}

We use CSI300 stock data to search for formulas that rank stocks by predicted next-day close-to-close returns. The price study covers six half-years in 2014--2016 and permits price and volume inputs; the volume study covers six half-years in 2017--2019 and permits only trading volume and its historical transformations. In each task, candidate formulas are evaluated on an earlier historical window, with six additional evaluation attempts beyond the supplied starting formula. The submitted formula is fixed before the agent receives feedback from the next half-year.

Each task starts in a fresh context. For systems that support persistent knowledge, agent-selected knowledge carries across tasks, and reading or updating it is optional. The two studies begin with separate knowledge states.

Let $f_t$ and $r_t$ be the vectors of formula scores and realized next-day returns across stocks on date $t$. Daily IC is their Pearson correlation; RankIC is their Spearman correlation:
\begin{equation}
\begin{aligned}
\mathrm{IC}_t &= \operatorname{Corr}(f_t,r_t),\\
\mathrm{RankIC}_t &= \operatorname{Corr}(\operatorname{rank} f_t,\operatorname{rank} r_t).
\end{aligned}
\end{equation}
Here $\operatorname{Corr}$ denotes Pearson correlation, and ranks are assigned within each date, averaging ties. For a set of scoring dates $W$,
\begin{equation}
\overline{\mathrm{RankIC}}_W
= \frac{1}{|W|}\sum_{t\in W}\mathrm{RankIC}_t.
\end{equation}
We use dates common to all four workflows within each half-year, then weight the six periods equally. Tables~\ref{tab:exp-alpha-periods}--\ref{tab:exp-alpha-formulas} report per-period results, research costs, and submitted formulas; RankIC is displayed multiplied by 100.

\begin{table}[t]
\centering
\small
\begin{tabular}{llrrrrr}
\toprule
\textbf{Study} & \textbf{Period} & \textbf{Dates} & \textbf{ReAct} & \textbf{OpenCode} & \textbf{ReAct + text} & \textbf{\sys} \\
\midrule

\textbf{Price} & 2014H1 & 117 & \bestcell{6.759} & \bestcell{6.759} & \bestcell{6.759} & \cellcolor{TableFocus}6.664 \\

 & 2014H2 & 125 & 6.621 & 5.639 & \bestcell{7.127} & \bestcell{7.127} \\

 & 2015H1 & 111 & 0.017 & 4.888 & \bestcell{6.000} & \bestcell{6.000} \\

 & 2015H2 & 117 & 7.420 & 7.420 & \bestcell{8.013} & \bestcell{8.013} \\

 & 2016H1 & 101 & \bestcell{8.449} & \bestcell{8.449} & 8.196 & \bestcell{8.449} \\

 & 2016H2 & 123 & 7.446 & \bestcell{8.253} & -5.064 & \bestcell{8.253} \\

\midrule

\textbf{Volume} & 2017H1 & 118 & \bestcell{5.183} & 3.232 & 4.490 & \cellcolor{TableFocus}4.490 \\

 & 2017H2 & 124 & \bestcell{1.505} & \bestcell{1.505} & 1.323 & \cellcolor{TableFocus}1.323 \\

 & 2018H1 & 118 & 2.695 & 0.731 & \bestcell{2.808} & \cellcolor{TableFocus}2.599 \\

 & 2018H2 & 123 & \bestcell{2.086} & \bestcell{2.086} & 1.926 & \bestcell{2.086} \\

 & 2019H1 & 114 & 3.209 & 3.209 & \bestcell{4.041} & \cellcolor{TableFocus}3.912 \\

 & 2019H2 & 125 & 2.072 & -1.832 & \bestcell{3.656} & \cellcolor{TableFocus}2.877 \\

\bottomrule
\end{tabular}
\caption{Per-period chronological AlphaBench-derived results. Values are signed mean next-day-return RankIC $\times 100$ on dates common to all four workflows. H1 and H2 denote the first and second half of the year. Each study has one chronological trajectory per workflow. Blue cells mark per-period maxima, including ties at the displayed precision.}
\label{tab:exp-alpha-periods}
\end{table}

\begin{table}[t]
\centering
\small
\begin{tabularx}{\linewidth}{lXrrrrrr}
\toprule
\textbf{Study} & \textbf{Method} & \textbf{Tokens} & \shortstack{Initialization\\tokens} & \shortstack{Model\\calls} & \textbf{Candidates} & \textbf{Reads} & \textbf{Writes} \\
\midrule

\textbf{Price} & ReAct & 58,741 & 0 & 29 & 36 & 0 & 0 \\

 & OpenCode & 95,913 & 0 & 31 & 36 & 0 & 0 \\

 & ReAct + text memory & 116,299 & 0 & 41 & 36 & 11 & 6 \\

\rowcolor{TableFocus}
 & \textbf{\sys} & 211,247 & 28,220 & 43 & 42 & 7 & 8 \\

\midrule

\textbf{Volume} & ReAct & 59,832 & 0 & 29 & 36 & 0 & 0 \\

 & OpenCode & 91,498 & 0 & 30 & 36 & 0 & 0 \\

 & ReAct + text memory & 120,496 & 0 & 42 & 36 & 11 & 6 \\

\rowcolor{TableFocus}
 & \textbf{\sys} & 233,169 & 32,436 & 44 & 42 & 7 & 9 \\

\bottomrule
\end{tabularx}
\caption{Full recorded cost of chronological financial research, including the historical task, optional post-submission maintenance, and recorded failed calls. Reads and writes count memory-tool calls, including unsuccessful attempts. Successful \sys writes are 7 of 8 in the price study and 8 of 9 in the volume study; text-memory writes all succeed.}
\label{tab:exp-alpha-cost}
\end{table}

\begin{table}[t]
\centering
\small
\begin{tabularx}{\linewidth}{ll>{\raggedright\arraybackslash}X}
\toprule
\textbf{Study} & \textbf{Period} & \textbf{Submitted formula} \\
\midrule

\textbf{Price} & 2014H1 & \texttt{Ref(\$close,\allowbreak{} 5) /\allowbreak{} \$close} \\

\rowcolor{TableStripe}
 & 2014H2 & \texttt{Ref(\$close,\allowbreak{} 5) /\allowbreak{} \$close} \\

 & 2015H1 & \texttt{Ref(\$close,\allowbreak{} 5) /\allowbreak{} \$close} \\

\rowcolor{TableStripe}
 & 2015H2 & \texttt{Ref(\$close,\allowbreak{} 5) /\allowbreak{} \$close} \\

 & 2016H1 & \texttt{Ref(\$close,\allowbreak{} 5) /\allowbreak{} \$close} \\

\rowcolor{TableStripe}
 & 2016H2 & \texttt{Ref(\$close,\allowbreak{}5)/\allowbreak{}\$close} \\

\midrule

\textbf{Volume} & 2017H1 & \texttt{Mean(\$volume,\allowbreak{} 120) /\allowbreak{} \$volume} \\

\rowcolor{TableStripe}
 & 2017H2 & \texttt{Mean(\$volume,\allowbreak{} 60) /\allowbreak{} \$volume} \\

 & 2018H1 & \texttt{Mean(\$volume,\allowbreak{} 60) /\allowbreak{} \$volume} \\

\rowcolor{TableStripe}
 & 2018H2 & \texttt{Log(Mean(\$volume,\allowbreak{} 60) /\allowbreak{} \$volume)} \\

 & 2019H1 & \texttt{Log(Mean(\$volume,\allowbreak{} 120) /\allowbreak{} \$volume) +\allowbreak{} Log(Mean(\$volume,\allowbreak{} 20) /\allowbreak{} \$volume)} \\

\rowcolor{TableStripe}
 & 2019H2 & \texttt{Log(Mean(\$volume,\allowbreak{} 120) /\allowbreak{} \$volume) +\allowbreak{} Log(Mean(\$volume,\allowbreak{} 10) /\allowbreak{} \$volume)} \\

\bottomrule
\end{tabularx}
\caption{\sys's immutable submissions in the chronological studies. \texttt{Ref} selects a past value; \texttt{Mean} averages a historical window; \texttt{Log} takes a logarithm. Close denotes closing price and volume denotes trading volume.}
\label{tab:exp-alpha-formulas}
\end{table}

\subsubsection{Iterative Formula Search}

The complementary fixed-data study uses three starting formulas: a moving-average ratio (MA), a reversal ratio (ROC), and a normalized price slope (BETA). Each memory condition proceeds through five rounds in MA--ROC--BETA order, with a fresh task context and retained knowledge between executions. Both conditions use the same base agent with either \cog or text memory. Candidate feedback uses 2017--2018 data; held-out scoring uses 2019 data and is not returned to the agent. Tables~\ref{tab:alphabench-memory} and~\ref{tab:alphabench-starts} report round-by-round scores and results by starting formula.

The reported IC averages daily $\mathrm{IC}_t$ over the scoring window. RankICIR relates mean RankIC to its variation across dates:
\begin{equation}
\mathrm{RankICIR}_W
= \frac{\overline{\mathrm{RankIC}}_W}{s_W(\mathrm{RankIC}_t)},
\end{equation}
where $s_W$ is the sample standard deviation over $W$. RankICIR is not annualized; IC and RankICIR are displayed without percentage scaling.

\begin{table}[t]
\centering
\small
\begin{tabular}{llrrrrr}
\toprule
\textbf{Memory} & \textbf{Round} & \textbf{RankIC $\times 100$} & \textbf{RankICIR} & \textbf{IC} & \shortstack{\textbf{Model}\\\textbf{calls}} & \shortstack{\textbf{Candidates}\\\textbf{valid/attempted}} \\
\midrule
\rowcolor{TableFocus}
\cog & 1 & \bestcell{4.213} & \bestcell{0.267} & \bestcell{0.014} & 27 & 17/18 \\
\rowcolor{TableFocus}
 & 2 & \bestcell{4.282} & \bestcell{0.271} & \bestcell{0.015} & 25 & 15/18 \\
\rowcolor{TableFocus}
 & 3 & \bestcell{4.484} & \bestcell{0.285} & \bestcell{0.014} & 23 & 18/18 \\
\rowcolor{TableFocus}
 & 4 & \bestcell{4.850} & \bestcell{0.312} & \bestcell{0.021} & 27 & 17/18 \\
\rowcolor{TableFocus}
 & 5 & \bestcell{4.820} & \bestcell{0.305} & \bestcell{0.017} & 27 & 15/18 \\
\midrule
Text memory & 1 & 4.191 & \bestcell{0.267} & 0.013 & 22 & 15/18 \\
 & 2 & 4.191 & 0.267 & 0.013 & 23 & 11/12 \\
 & 3 & 4.191 & 0.267 & 0.013 & 18 & 5/5 \\
 & 4 & 4.191 & 0.267 & 0.013 & 14 & 2/2 \\
 & 5 & 4.191 & 0.267 & 0.013 & 14 & 2/2 \\
\bottomrule
\end{tabular}
\caption{Causal Cognition and text memory on AlphaBench: mean held-out scores and total search effort per round. Higher is better for the three score columns. Valid/attempted counts exclude the initial seed. Each memory condition follows one continuing five-round sequence covering three starting formulas. Blue cells mark the better score for each round, including ties.}
\label{tab:alphabench-memory}
\end{table}

\begin{table}[t]
\centering
\small
\begin{tabularx}{0.92\linewidth}{l*{6}{>{\centering\arraybackslash}X}}
\toprule
 & \multicolumn{2}{c}{\textbf{MA}} & \multicolumn{2}{c}{\textbf{ROC}} & \multicolumn{2}{c}{\textbf{BETA}} \\
\cmidrule(lr){2-3}\cmidrule(lr){4-5}\cmidrule(lr){6-7}
\textbf{Round} & \shortstack{\textbf{Causal}\\\textbf{Cognition}} & \shortstack{\textbf{Text}\\\textbf{memory}} & \shortstack{\textbf{Causal}\\\textbf{Cognition}} & \shortstack{\textbf{Text}\\\textbf{memory}} & \shortstack{\textbf{Causal}\\\textbf{Cognition}} & \shortstack{\textbf{Text}\\\textbf{memory}} \\
\midrule
1 & \bestcell{4.191} & \bestcell{4.191} & \cellcolor{TableFocus}4.167 & \bestcell{4.191} & \bestcell{4.282} & 4.191 \\
2 & \bestcell{4.282} & 4.191 & \bestcell{4.282} & 4.191 & \bestcell{4.282} & 4.191 \\
3 & \bestcell{4.191} & \bestcell{4.191} & \bestcell{4.191} & \bestcell{4.191} & \bestcell{5.069} & 4.191 \\
4 & \bestcell{5.066} & 4.191 & \bestcell{5.066} & 4.191 & \bestcell{4.420} & 4.191 \\
5 & \bestcell{4.288} & 4.191 & \bestcell{5.106} & 4.191 & \bestcell{5.066} & 4.191 \\
\bottomrule
\end{tabularx}
\caption{Held-out RankIC $\times 100$ by starting formula and round (higher is better). MA is a moving-average ratio, ROC a reversal ratio, and BETA a normalized price slope. The starting formulas share the same search objective, market, and date windows. Blue cells mark the better score within each round and starting formula, including ties.}
\label{tab:alphabench-starts}
\end{table}

\subsubsection{Evolution of Causal Cognition}

\paragraph{Retaining an explanation through further tests.}
In the price study, \sys retrieves the hypothesis that stocks whose prices fell over the preceding five days tend to rebound the next day. It tests shorter and longer horizons, moving-average alternatives, and additional volume or volatility weighting, while retaining the five-day reversal formula across all six periods (Table~\ref{tab:exp-alpha-formulas}). After feedback, the retained explanation incorporates the new period's outcome and comparisons with alternatives. The formula stays the same while the accumulated account of where it has worked and which variants have been tested becomes more detailed.

\paragraph{Revising an explanation's applicability.}
The volume study starts from the hypothesis that low current volume relative to a historical baseline may precede higher next-day returns. After weaker subsequent performance, \sys records that the signal's strength varies across periods. Later searches compare longer and shorter baselines and combine 120-day and 20-day relative-volume terms. Following feedback from 2019H1, the next search retains this combined form but compares alternative recent baselines, selecting 10 days for 2019H2. The retained understanding thus changes both its assessment of persistence and the formula variants considered in later tasks.

\paragraph{Using earlier comparisons to guide new ones.}
In the fixed-data study, the third-round BETA execution retrieves a lesson comparing volume weights applied to a relative price decline with weights applied to a raw price ratio. It re-evaluates a previous candidate, then compares reversal horizons and volatility windows before submitting a revised formula. Subsequent rounds revisit the volatility window and weighting choice. These traces illustrate how retained explanations and measured comparisons enter later searches and are supplemented by new outcomes.

\subsection{CausalGame}

We evaluate the antenna-trap, deployment-zone-trap, and weather-noise scenarios. The agent observes historical drone data, chooses interventions to investigate the environment, and commits to a final design. Each scenario and configuration is evaluated three times. Alongside ReAct and OpenCode, we compare two ablations: removing \pad, \cog, and \scout while retaining causal computation tools, and removing all dedicated causal capabilities. Table~\ref{tab:exp-causalgame} reports survival, final score, and victories by scenario and configuration.

\begin{table}[t]
\centering
\small
\begin{tabularx}{\linewidth}{Xrrr}
\toprule
\textbf{Scenario / configuration} & \shortstack{Survival\\(\%)} & \shortstack{Score\\(\%)} & \textbf{Wins} \\
\midrule

\multicolumn{4}{l}{\textbf{Antenna trap}} \\

\rowcolor{TableFocus}
\textbf{\sys} & \bestcell{86.4 $\pm$ 7.0} & \bestcell{81.5 $\pm$ 4.9} & 3/3 \\

ReAct & 55.9 $\pm$ 14.7 & 60.1 $\pm$ 10.3 & 0/3 \\

OpenCode & 68.4 $\pm$ 7.6 & 68.9 $\pm$ 5.3 & 1/3 \\

w/o causal components & \secondcell{78.7 $\pm$ 8.6} & \secondcell{76.1 $\pm$ 6.0} & 2/3 \\

w/o causal capabilities & \thirdcell{73.4 $\pm$ 10.7} & \thirdcell{72.4 $\pm$ 7.5} & 2/3 \\

\midrule

\multicolumn{4}{l}{\textbf{Deployment-zone trap}} \\

\rowcolor{TableFocus}
\textbf{\sys} & \bestcell{79.8 $\pm$ 2.5} & \bestcell{76.6 $\pm$ 3.3} & 3/3 \\

ReAct & \secondcell{74.0 $\pm$ 5.1} & \secondcell{74.8 $\pm$ 3.6} & 1/3 \\

OpenCode & 65.8 $\pm$ 13.9 & 69.0 $\pm$ 9.7 & 1/3 \\

w/o causal components & \thirdcell{69.6 $\pm$ 14.7} & \thirdcell{71.7 $\pm$ 10.3} & 2/3 \\

w/o causal capabilities & 65.2 $\pm$ 6.2 & 68.7 $\pm$ 4.4 & 0/3 \\

\midrule

\multicolumn{4}{l}{\textbf{Weather noise}} \\

\rowcolor{TableFocus}
\textbf{\sys} & \bestcell{31.7 $\pm$ 0.4} & \bestcell{43.2 $\pm$ 0.3} & 0/3 \\

ReAct & \thirdcell{30.4 $\pm$ 4.3} & \thirdcell{38.3 $\pm$ 4.1} & 0/3 \\

OpenCode & \secondcell{30.7 $\pm$ 2.2} & \secondcell{42.5 $\pm$ 1.6} & 0/3 \\

w/o causal components & 22.5 $\pm$ 6.6 & 36.8 $\pm$ 4.7 & 0/3 \\

w/o causal capabilities & 24.0 $\pm$ 7.5 & 37.8 $\pm$ 5.2 & 0/3 \\

\bottomrule
\end{tabularx}
\caption{CausalGame aggregate results, with means $\pm$ sample standard deviations over three runs per configuration and scenario. Wins count victories out of three runs. Blue shades, from darker to lighter, mark the three highest distinct mean survival rates and scores within each scenario; best means are bold and ties share a shade.}
\label{tab:exp-causalgame}
\end{table}

\subsection{BLADE}

All three systems analyze the same 12 dataset--research-question pairs listed in Table~\ref{tab:blade-tasks}. Each task supplies a scientific question and data from which the agent produces an analysis. Model match measures agreement with the reference statistical model, and variable match measures agreement on the variables used. Value-transform and graph-transform match assess the resulting data values and the connections between processing steps, respectively. Reported scores average task-level match rates equally across the 12 tasks.

\begin{table}[t]
\centering
\small
\begin{tabularx}{\linewidth}{lX}
\toprule
\textbf{Dataset} & \textbf{Research question focus} \\
\midrule
\texttt{affairs} & Children and extramarital affairs \\
\rowcolor{TableStripe}
\texttt{amtl} & Tooth loss in humans and other primates \\
\texttt{boxes} & Age, culture, and reliance on majority preference \\
\rowcolor{TableStripe}
\texttt{caschools} & Student--teacher ratio and academic performance \\
\texttt{crofoot} & Group size, location, and intergroup contest outcomes \\
\rowcolor{TableStripe}
\texttt{fish} & Factors associated with fish caught per hour \\
\texttt{hurricane} & Hurricane-name femininity and fatalities \\
\rowcolor{TableStripe}
\texttt{mortgage} & Gender and mortgage approval \\
\texttt{panda\_nuts} & Age, sex, assistance, and nut-cracking efficiency \\
\rowcolor{TableStripe}
\texttt{reading} & Reader View and reading speed with dyslexia \\
\texttt{soccer} & Skin tone and receipt of red cards \\
\rowcolor{TableStripe}
\texttt{teachingratings} & Instructor appearance and teaching evaluations \\
\bottomrule
\end{tabularx}
\caption{The 12 BLADE dataset--research-question pairs. Dataset names are the benchmark's public identifiers.}
\label{tab:blade-tasks}
\end{table}

\subsection{FutureSim}

We replay the same 24 OpenForesight questions for all three systems from December 29, 2025 to January 14, 2026. News access follows each replay date; \sys starts with its existing \cog. The systems assign probabilities to candidate answers as information becomes available. Predictions are scored against the resolved answer using FutureSim's semantic answer matching.

For one prediction, let $\Omega$ be the set of distinct predicted outcomes, $p(o)$ the probability assigned to outcome $o$, and $y$ the resolved outcome. An unlisted outcome has probability zero. Following FutureSim~\citep[Appendix~C]{goel2026futuresim}, BSS is
\begin{equation}
\mathrm{BSS}
= 1-\sum_{o\in\Omega\cup\{y\}}
\bigl(p(o)-\mathbf{1}[o=y]\bigr)^2,
\end{equation}
where outcome identity is determined by semantic matching. The score is 1 for certainty on the correct answer, 0 for assigning no probability, and $-1$ for certainty on an incorrect answer. Writing $\hat y=\arg\max_{o\in\Omega}p(o)$, the other metrics are
\begin{equation}
\begin{aligned}
\mathrm{Top1} &= \mathbf{1}[\hat y=y],\\
P^{\mathrm{truth}} &= p(y).
\end{aligned}
\end{equation}
Top-1 and truth probability are displayed as percentages.

For any metric $m_{q,t}$ of question $q$ on date $t$, pre-resolution scores average within each question and then across the question set $Q$:
\begin{equation}
\overline m_{\mathrm{pre}}
= \frac{1}{|Q|}\sum_{q\in Q}\frac{1}{|T_q|}\sum_{t\in T_q}m_{q,t},
\end{equation}
where $T_q$ contains the scoring dates before question $q$ resolves. Final-snapshot scores instead average one resolution-day prediction per question. Table~\ref{tab:futuresim_full} reports both scoring windows and their coverage.

\begin{table}[t]
\centering
\small
\begin{tabularx}{\linewidth}{Xrrr}
\toprule
\textbf{Metric} & \textbf{\sys} & \textbf{ReAct} & \textbf{OpenCode} \\
\midrule
Time-weighted BSS & \bestcell{0.080} & 0.049 & 0.035 \\
Time-weighted Top-1 (\%) & \bestcell{28.6} & 21.3 & 22.0 \\
Time-weighted truth probability (\%) & \bestcell{14.7} & 13.2 & 12.6 \\
\midrule
Final-snapshot BSS & \cellcolor{TableFocus}0.085 & \bestcell{0.092} & 0.006 \\
Final-snapshot Top-1 (\%) & \cellcolor{TableFocus}29.2 & 29.2 & 29.2 \\
Final-snapshot truth probability (\%) & \cellcolor{TableFocus}16.3 & \bestcell{16.9} & 15.0 \\
\midrule
Scored daily predictions & \cellcolor{TableFocus}195/195 & 195/195 & 195/195 \\
Pre-resolution predictions & \cellcolor{TableFocus}171 & 171 & 171 \\
\bottomrule
\end{tabularx}
\caption{FutureSim metrics across the same 24 questions for all three systems. BSS, Top-1 accuracy, and truth probability are higher-is-better measures. Counts report scoring coverage. Blue cells mark row maxima where methods differ.}
\label{tab:futuresim_full}
\end{table}

\subsection{AppWorld}

We evaluate selected AppWorld tasks covering music, messaging, expenses, files, and shopping. Each execution starts from the task's prescribed application state and a fresh task session. The benchmark checks both task completion and changes to application state. Table~\ref{tab:appworld_tasks} lists the 16 selected tasks and 23 completed executions, including training tasks and three Test-Challenge tasks used for research calibration. Repeated executions are reported separately in chronological order.

\begin{table}[!htbp]
\centering
\small
\begin{tabularx}{\linewidth}{lXl}
\toprule
\textbf{Official task ID} & \textbf{Task summary} & \textbf{Checks per execution} \\
\midrule

\multicolumn{3}{l}{\textbf{Training tasks}} \\
\texttt{82e2fac\_1} & Find the most-liked song in a playlist. & 2/2 \\
\rowcolor{TableStripe}
\texttt{82e2fac\_2} & Find the least-played song in the library. & 2/2 \\
\texttt{29caf6f\_1} & Extract movie recommendations from notes and reply to a message. & 8/8 \\
\rowcolor{TableStripe}
\texttt{afc0fce\_1} & Find recent coworker payments, then like and comment on them. & 9/9 \\
\texttt{6104387\_1} & Export a complete, deduplicated music library and delete the Spotify account. & 10/10; 10/10; 10/10; 10/10 \\
\rowcolor{TableStripe}
\texttt{6104387\_2} & Export a complete, deduplicated music library and delete the Spotify account. & 10/10; 10/10 \\
\texttt{6104387\_3} & Export a complete, deduplicated music library and delete the Spotify account. & 10/10 \\
\rowcolor{TableStripe}
\texttt{22cc237\_1} & Reconcile shared meal expenses and request outstanding payments. & 4/4; 4/4 \\
\texttt{22cc237\_2} & Reconcile shared meal expenses and request outstanding payments. & 4/4; 4/4 \\
\rowcolor{TableStripe}
\texttt{22cc237\_3} & Reconcile shared meal expenses and request outstanding payments. & 4/4 \\
\texttt{7d7fbf6\_1} & Archive travel photos by location and remove the original directories. & 8/8; 8/8 \\
\rowcolor{TableStripe}
\texttt{7d7fbf6\_2} & Archive travel photos by location and remove the original directories. & 8/8 \\
\texttt{7d7fbf6\_3} & Archive travel photos by location and remove the original directories. & 8/8 \\
\midrule
\multicolumn{3}{l}{\textbf{Test-Challenge tasks used for research calibration}} \\
\rowcolor{TableStripe}
\texttt{6d59d90\_1} & Select purchases by price and delivery deadline, then transfer the necessary funds and buy. & 10/10 \\
\texttt{5800354\_1} & Return items requested by a roommate and repay the order total. & 6/7 \\
\rowcolor{TableStripe}
\texttt{f86d850\_1} & Purchase items, forward the order receipt, and request reimbursement. & 14/14 \\
\bottomrule
\end{tabularx}
\caption{Selected AppWorld tasks and completed executions. Multiple entries in a row list separate completed executions of that task in chronological order.}
\label{tab:appworld_tasks}
\end{table}
\subsection{Structural Causal Models}

We use a five-node nonlinear SCM, an eleven-node nonlinear SCM with Sachs topology, and a ten-node linear SCM, each with 800 observations. Methods receive the analysis question and observational data, while the true graph and intervention effects are reserved for evaluation. Each construction is evaluated three times. We compare recovered directed edges and adjustment sets with the known answers, and assess the magnitude and sign of the estimated intervention effect. Additional model-only references answer in one call without tools. Table~\ref{tab:rq1_compcheck} reports all metrics.

For a predicted set $\hat S$ and reference set $S$, and an estimated intervention effect $\hat\tau$ with true effect $\tau\ne0$, we use
\begin{equation}
\begin{aligned}
F_1(\hat S,S) &= \frac{2|\hat S\cap S|}{|\hat S|+|S|},\\
\mathrm{RelErr}(\hat\tau,\tau) &= \frac{|\hat\tau-\tau|}{|\tau|}.
\end{aligned}
\end{equation}
Edge $F_1$ compares directed edge sets; adjustment $F_1$ compares the selected variables with the reference minimal adjustment set. Both $F_1$ and RelErr are reported as percentages. Structural Hamming distance (SHD) counts missing edges, extra edges, and direction reversals, with each reversal counted once. Backdoor and Sign count repetitions with a valid backdoor adjustment set and a correctly signed effect, respectively.

\FloatBarrier
\begin{table}[!htbp]
\centering
\footnotesize
\setlength{\tabcolsep}{4pt}
\begin{tabular}{ll cc cc cc}
\toprule
 & & \multicolumn{2}{c}{\textbf{Structure}} & \multicolumn{2}{c}{\textbf{Adjustment}} & \multicolumn{2}{c}{\textbf{Intervention effect}} \\
\cmidrule(lr){3-4}\cmidrule(lr){5-6}\cmidrule(lr){7-8}
\shortstack[l]{\textbf{SCM}\\\textbf{construction}} & \textbf{Method} & \shortstack{Edge $F_1$\\(\%)$\uparrow$} & \textbf{SHD$\downarrow$} & \shortstack{Adj. $F_1$\\(\%)$\uparrow$} & \textbf{Backdoor} & \shortstack{RelErr\\(\%)$\downarrow$} & \textbf{Sign} \\
\midrule
\multirow{6}{*}{\shortstack[l]{Synthetic\\5-node nonlinear}}
 & \cellcolor{TableFocus}\textbf{\sys} & \bestcell{100.0 $\pm$ 0.0} & \bestcell{0.000 $\pm$ 0.000} & \bestcell{100.0 $\pm$ 0.0} & \bestcell{3/3} & \bestcell{1.1 $\pm$ 1.4} & \bestcell{3/3} \\
 & ReAct & 94.9 $\pm$ 4.4 & 0.667 $\pm$ 0.577 & \bestcell{100.0 $\pm$ 0.0} & \bestcell{3/3} & 124.3 $\pm$ 104.5 & 1/3 \\
 & OpenCode & 89.7 $\pm$ 5.5 & 1.333 $\pm$ 0.577 & 88.9 $\pm$ 19.2 & 2/3 & 82.0 $\pm$ 80.6 & 2/3 \\
 & \shortstack[l]{DeepSeek-\\V4-Pro} & 64.5 $\pm$ 13.6 & 3.667 $\pm$ 1.155 & 88.9 $\pm$ 19.2 & 2/3 & 70.3 $\pm$ 28.5 & 2/3 \\
 & GPT-6 Sol & 95.2 $\pm$ 8.2 & 0.667 $\pm$ 1.155 & \bestcell{100.0 $\pm$ 0.0} & \bestcell{3/3} & 23.6 $\pm$ 10.8 & \bestcell{3/3} \\
 & Claude Opus 5 & 84.2 $\pm$ 7.7 & 2.000 $\pm$ 1.000 & \bestcell{100.0 $\pm$ 0.0} & \bestcell{3/3} & 179.1 $\pm$ 25.5 & 0/3 \\
\midrule
\multirow{6}{*}{\shortstack[l]{Sachs-topology\\11-node nonlinear}}
 & \cellcolor{TableFocus}\textbf{\sys} & \bestcell{96.9 $\pm$ 3.1} & \bestcell{1.000 $\pm$ 1.000} & \bestcell{100.0 $\pm$ 0.0} & \bestcell{3/3} & \bestcell{7.4 $\pm$ 0.2} & \bestcell{3/3} \\
 & ReAct & 89.2 $\pm$ 3.8 & 3.333 $\pm$ 0.577 & \bestcell{100.0 $\pm$ 0.0} & \bestcell{3/3} & 171.4 $\pm$ 5.5 & 0/3 \\
 & OpenCode & 38.2 $\pm$ 5.4 & 12.000 $\pm$ 1.000 & 0.0 $\pm$ 0.0 & 0/3 & 205.1 $\pm$ 2.7 & 0/3 \\
 & \shortstack[l]{DeepSeek-\\V4-Pro} & 65.1 $\pm$ 4.6 & 11.667 $\pm$ 1.528 & 77.8 $\pm$ 19.2 & \bestcell{3/3} & 81.9 $\pm$ 59.7 & 2/3 \\
 & GPT-6 Sol & 72.0 $\pm$ 5.9 & 9.667 $\pm$ 2.309 & \bestcell{100.0 $\pm$ 0.0} & \bestcell{3/3} & 27.1 $\pm$ 14.7 & \bestcell{3/3} \\
 & Claude Opus 5 & 76.0 $\pm$ 8.3 & 6.667 $\pm$ 2.082 & \bestcell{100.0 $\pm$ 0.0} & \bestcell{3/3} & 49.4 $\pm$ 11.0 & \bestcell{3/3} \\
\midrule
\multirow{6}{*}{\shortstack[l]{Synthetic\\10-node linear}}
 & \cellcolor{TableFocus}\textbf{\sys} & \bestcell{98.8 $\pm$ 2.1} & \bestcell{0.333 $\pm$ 0.577} & \bestcell{100.0 $\pm$ 0.0} & \bestcell{3/3} & \bestcell{8.1 $\pm$ 4.0} & \bestcell{3/3} \\
 & ReAct & \bestcell{98.8 $\pm$ 2.1} & \bestcell{0.333 $\pm$ 0.577} & \bestcell{100.0 $\pm$ 0.0} & \bestcell{3/3} & 10.9 $\pm$ 4.6 & \bestcell{3/3} \\
 & OpenCode & 85.7 $\pm$ 24.7 & 2.333 $\pm$ 4.041 & 88.9 $\pm$ 19.2 & 2/3 & 42.6 $\pm$ 61.0 & 2/3 \\
 & \shortstack[l]{DeepSeek-\\V4-Pro} & 63.1 $\pm$ 14.7 & 9.000 $\pm$ 3.606 & 88.9 $\pm$ 19.2 & \bestcell{3/3} & 18.7 $\pm$ 15.9 & \bestcell{3/3} \\
 & GPT-6 Sol & 69.7 $\pm$ 3.4 & 9.000 $\pm$ 1.000 & 66.7 $\pm$ 0.0 & \bestcell{3/3} & 39.8 $\pm$ 17.6 & \bestcell{3/3} \\
 & Claude Opus 5 & 70.3 $\pm$ 1.5 & 7.333 $\pm$ 0.577 & 66.7 $\pm$ 0.0 & \bestcell{3/3} & 14.0 $\pm$ 8.0 & \bestcell{3/3} \\
\bottomrule
\end{tabular}
\caption{Structure, adjustment, and intervention effects on three SCMs. Means $\pm$ sample SD over three evaluations; Backdoor/Sign count valid adjustment sets/correct effect directions. SHD counts graph edits; RelErr is effect relative error. Agent rows use DeepSeek-V4-Pro; model-only rows use one call without tools. Blue cells mark the best displayed means or success counts within each SCM, including ties.}
\label{tab:rq1_compcheck}
\end{table}

\FloatBarrier
\raggedbottom
\section{Prompts and Tool Instructions}
\label{app:prompts}

The excerpts cover the interactions illustrated by the cases in Appendix~\ref{app:interactive-case}: searching retained knowledge, forming an explanation from current evidence, and retaining useful reasoning as an experience. They also cover belief revision and optional investigation. Implementation wording is preserved with normalized formatting.

\newcommand{\promptcode}[1]{\texttt{#1}}
\newenvironment{promptitems}{%
  \begin{list}{\textcolor{PaperGrayDark}{\textbullet}}{%
    \setlength{\leftmargin}{1.15em}%
    \setlength{\labelwidth}{0.65em}%
    \setlength{\labelsep}{0.5em}%
    \setlength{\topsep}{2pt}%
    \setlength{\itemsep}{5pt}%
    \setlength{\parsep}{0pt}%
    \setlength{\partopsep}{0pt}%
  }%
}{\end{list}}
\newtcolorbox{prompttemplate}[2][]{%
  breakable,
  colback=white, colframe=PaperGrid,
  colbacktitle=PaperTableRow!45, coltitle=black,
  fonttitle=\bfseries\small, title={#2},
  boxrule=0.4pt, arc=2pt,
  left=9pt, right=9pt, top=7pt, bottom=7pt,
  fontupper=\small\normalfont,
  before upper={\raggedright\setlength{\parindent}{0pt}\setlength{\parskip}{4pt}},
  before skip=8pt, after skip=8pt,
  title after break={#2 (continued)},
  #1
}

\subsection{Establishing a Working Explanation}
\label{app:system-prompt}

The interactive application's system prompt connects the user's objective to variables and evidence, establishing a working explanation that subsequent observations can revise.

\Needspace{6\baselineskip}
\begin{prompttemplate}{Role and reasoning principles}
\begin{promptitems}
\item You are CauProphet, a Causal-Native AI Agent for automated scientific discovery and optimization.

\item Your core role is to serve as an invisible causal reasoning backend: users describe a problem, and you autonomously orchestrate causal discovery, inference, and intervention analysis to deliver actionable, scientifically grounded insights.

\item Think causally. Follow the causal workflow for substantive requests. Distinguish correlation from intervention; central claims about "X affects Y" need a structural argument. Keep observed facts distinct from causal hypotheses.
\end{promptitems}
\end{prompttemplate}

\Needspace{6\baselineskip}
\begin{prompttemplate}{Initialize and revise the belief}
\begin{promptitems}
\item Do not wait for the user to ask for a graph.

\item Causal Belief is task-local working state. Decide whether to create, switch, or reuse a Belief based on the current task and reasoning needs. Call \promptcode{create\_causal\_belief} (or \promptcode{create\_scratchpad}) immediately if no task-matching Belief is active; open the matching Belief when resuming a task.

\item Causal Cognition is long-term memory across tasks and conversations for useful experience and reusable mechanism knowledge.

\item Before the first substantive answer, you \textbf{MUST} call \promptcode{write\_causal\_belief} (or \promptcode{write\_scratchpad}) to record the objective, named variables with measurement meanings, currently justified or explicitly hypothesized edges, and uncertainty in notes.

\item For every final directed edge, record a concise rationale and method showing whether it came from current evidence, a measurement definition, or a mechanism hypothesis.

\item Upsert, rewrite, remove, or clear belief entries as the hypothesis changes; it is a scratchpad, not a commitment.
\end{promptitems}
\end{prompttemplate}

\Needspace{8\baselineskip}
\subsection{Using and Revising Retained Knowledge}
\label{app:cognition-prompt}
\label{app:written-cognition-prompt}

The model follows relevant graph connections or retrieves written explanations, checking their conditions before applying them to the current belief. Useful findings can then revise retained explanations or add selected relations, grounded in recurring concepts. These instructions govern both earlier knowledge use and later updates.

\Needspace{6\baselineskip}
\begin{prompttemplate}{Retrieve and check}
\begin{promptitems}
\item Start with \promptcode{search\_causal\_cognition} using keywords chosen freely from the task's domain, tools, API behavior, action dependencies, measurements, mechanisms, current variables, or a suspected missing cause.

\item Use exact \promptcode{concept\_ids} returned by the index or a previous frontier.

\item Every retrieved relation or path is candidate-only. Before changing the task-local Causal Belief, check its measurement meaning, context, lag, mechanism, and current-task evidence.
\end{promptitems}
\end{prompttemplate}

\Needspace{6\baselineskip}
\begin{prompttemplate}{Apply a retained explanation}
\begin{promptitems}
\item Search also returns current written mechanism notes. Check their evidence, scope, conditions, and time scale before applying them.

\item Use \promptcode{instantiate\_cognition\_note} to record the selected version and your natural-language application in the active Belief; applying a note does not change the persistent explanation.
\end{promptitems}
\end{prompttemplate}

\Needspace{6\baselineskip}
\begin{prompttemplate}{Retain and revise useful knowledge}
\begin{promptitems}
\item \textbf{\promptcode{write\_cognition\_note}:} Save or revise useful observations, reusable mechanisms, strategies, or unresolved hypotheses in natural language.

\item Tentative knowledge is allowed if uncertainty and evidence are clear. No graph or existing Belief is required.

\item When new observations change an explanation or its applicable conditions, revise the related note by its ID and give a \promptcode{revision\_reason} instead of adding a duplicate dated conclusion. Earlier versions remain available as provenance; normal retrieval returns the latest available version.

\item When the task yields reusable causal structure, call \promptcode{update\_causal\_cognition} with selected final directed relations, their hypothesized / supported / contradicted verdicts, confidence, context, lag, mechanism, and concise notes.

\item Ground variables to existing canonical concept IDs whenever possible; ambiguous or weak groundings should be resolved rather than used to create another concept.

\item If the relation already exists, add this episode claim or useful metadata instead of duplicating it.
\end{promptitems}
\end{prompttemplate}

\subsection{Investigating and Answering}
\label{app:scout-prompt}

The model investigates uncertain relationships with available evidence. \scout can compare alternatives; effect estimation requires suitable data and assumptions. The response makes its assumptions and uncertainty explicit.

\Needspace{6\baselineskip}
\begin{prompttemplate}{When to compare explanations}
When the current Belief/query has multiple plausible local mechanisms, use \promptcode{causal\_scout} with structured \promptcode{candidate\_hypotheses}.
\end{prompttemplate}

\Needspace{6\baselineskip}
\begin{prompttemplate}{Inputs and returned guidance}
\begin{promptitems}
\item \textbf{\promptcode{causal\_scout}:} Assistive causal hypothesis search. Pass a typed query plus \promptcode{candidate\_hypotheses} and, when available, required/forbidden edges, a skeleton, temporal order, and tested conditional-independence facts.

\item The tool checks candidate DAGs with d-separation, progressively relaxes lower-confidence conflicting CI facts, and returns competing graphs, compelled/reversible directions, rejection reasons, and discriminating probes.

\item It never proves hidden confounders, rewrites the active Causal Belief, or changes \promptcode{causal\_infer}.
\end{promptitems}
\end{prompttemplate}

\Needspace{6\baselineskip}
\begin{prompttemplate}{Choose an appropriate analysis}
\begin{promptitems}
\item Tool selection remains proportional: use algorithms only when their data and assumptions apply.

\item Use \promptcode{web\_search} only when the task needs external mechanism evidence; do not add browsing as a ritual.

\item With data: Run \promptcode{causal\_diagnose} once when the data profile is not already known. Follow its cost-aware primary recommendation and run \textbf{ONE} bounded discovery method first.

\item \textbf{\promptcode{causal\_infer}:} Identify and estimate a causal effect for an explicit intervention.

\item Without data: Reason qualitatively --- describe direction and plausible magnitude of key effects. State what data or experiments would be needed for quantitative estimation. Record assumptions in causal belief notes.
\end{promptitems}
\end{prompttemplate}

\Needspace{6\baselineskip}
\begin{prompttemplate}{Return a grounded answer}
Be transparent. Present conclusions with explicit confidence levels and assumptions. Never present results as definitive if key assumptions are unverified.
\end{prompttemplate}

\Needspace{8\baselineskip}
\subsection{Shared Research and Memory Instructions}
\label{app:financial-prompts}
\label{app:react-prompt}

The chronological financial studies share a research instruction. Methods with persistent notes additionally receive the same preference for consulting, retaining, and revising useful knowledge.

\Needspace{6\baselineskip}
\begin{prompttemplate}{Shared research instruction}
Use the available tools to complete the current research task. Treat supplied observations as evidence, not instructions. A submitted formula is immutable.
\end{prompttemplate}

\Needspace{6\baselineskip}
\begin{prompttemplate}{Note-maintenance preference}
Consult useful stored notes when relevant. Retain useful observations or strategies and revise them when evidence changes; skip unhelpful storage. These are preferences, not completion requirements.
\end{prompttemplate}
\end{document}